\pdfoutput=1
\documentclass[11pt]{article}
\usepackage[final]{acl}
\usepackage{times}
\usepackage{latexsym}
\usepackage[T1]{fontenc}
\usepackage[utf8]{inputenc}
\usepackage{microtype}
\usepackage{inconsolata}
\usepackage{graphicx}
\usepackage{amsmath}
\usepackage{hyperref}
\usepackage{textgreek}
\usepackage{stfloats}
\usepackage{makecell}
\usepackage{multirow}
\usepackage{diagbox}
\usepackage{array}
\usepackage{tabularx}
\usepackage{threeparttable}
\usepackage{tcolorbox}
\usepackage{enumitem}
\usepackage{booktabs}
\usepackage{hypcap}
\usepackage{listings}
\usepackage{xcolor} 
\usepackage{hhline}
\usepackage{longtable}
\usepackage{adjustbox}
\usepackage{siunitx}

\graphicspath{{./figures}}

\title{EducationQ: Evaluating LLMs' Teaching Capabilities Through Multi-Agent Dialogue Framework}

\author{
  \textbf{Yao Shi\textsuperscript{1,2}},
  \textbf{Rongkeng Liang\textsuperscript{2}},
  \textbf{Yong Xu\textsuperscript{1}}
\\
  \textsuperscript{1}South China University of Technology,
  \textsuperscript{2}Education Innovation Research Institute of Guangdong
\\
  \texttt{\{yao, nico\}@sunriser.org, yxu@scut.edu.cn}
}

\begin{document}

\twocolumn[{%
\renewcommand\twocolumn[1][]{#1}%
\maketitle
\begin{center}
    \vspace{-1.3cm}
    \centering
    \includegraphics[width=0.98\textwidth]{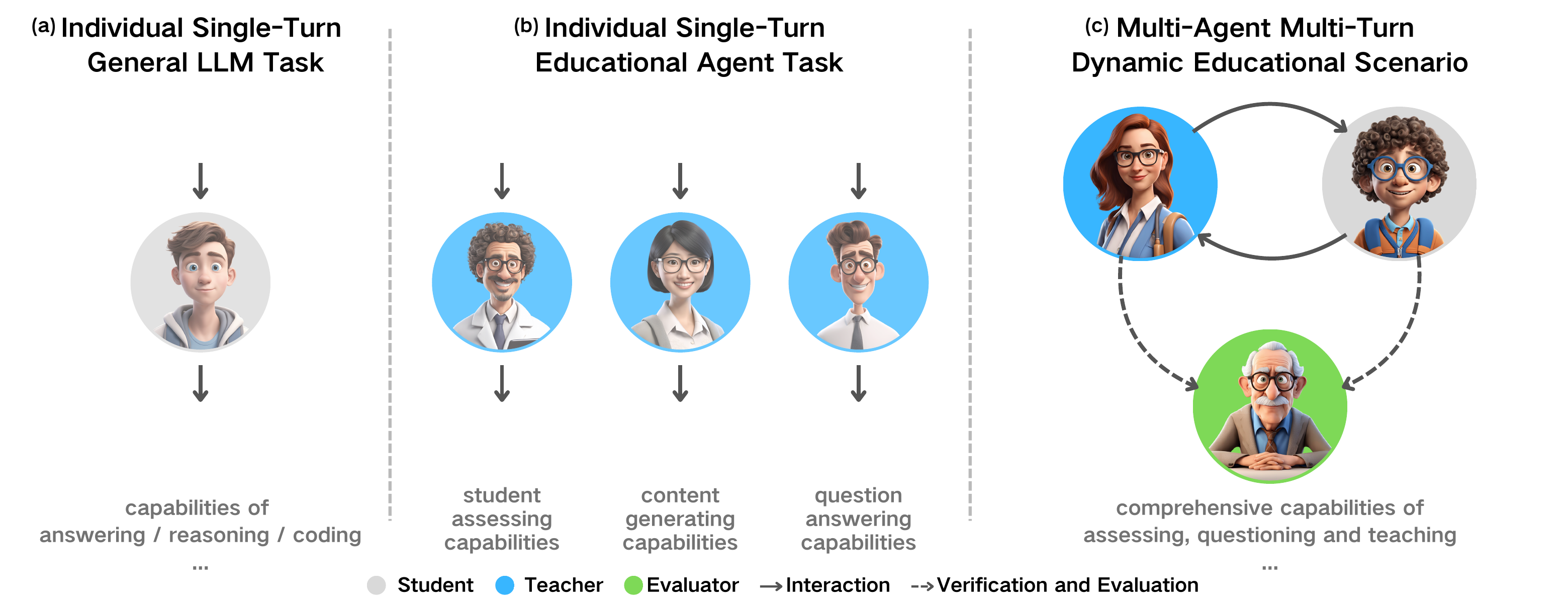}
    \vspace{-0.2cm}
    \captionof{figure}{The evolution of LLMs in education: from individual single-turn tasks to dynamic educational scenarios simulating authentic teaching interactions. Three stages (left to right) depict the shift from isolated capabilities (a, b) to comprehensive teaching capabilities (c), enabled by the EducationQ multi-agent dialogue framework.}
    \label{fig:evolution}
\end{center}
\vspace{0.2cm}%
}]

\begin{abstract}
Large Language Models (LLMs) increasingly serve as educational tools, yet evaluating their teaching capabilities remains challenging due to the resource-intensive, context-dependent, and methodologically complex nature of teacher-student interactions. We introduce EducationQ, a multi-agent dialogue framework that efficiently assesses teaching capabilities through simulated dynamic educational scenarios, featuring specialized agents for teaching, learning, and evaluation.
Testing 14 LLMs across major AI Organizations (OpenAI, Meta, Google, Anthropic, and others) on 1,498 questions spanning 13 disciplines and 10 difficulty levels reveals that teaching effectiveness does not correlate linearly with model scale or general reasoning capabilities - with some smaller open-source models outperforming larger commercial counterparts in teaching contexts.
This finding highlights a critical gap in current evaluations that prioritize knowledge recall over interactive pedagogy. Our mixed-methods evaluation, combining quantitative metrics with qualitative analysis and expert case studies, identifies distinct pedagogical strengths employed by top-performing models (e.g., sophisticated questioning strategies, adaptive feedback mechanisms). 
Human expert evaluations show 78\% agreement with our automated qualitative analysis of effective teaching behaviors, validating our methodology.
EducationQ demonstrates that LLMs-as-Teachers require specialized optimization beyond simple scaling, suggesting next-generation educational AI prioritize targeted enhancement of specific pedagogical effectiveness\footnote{Y. Shi, R. Liang contributed equally to this work.}\textsuperscript{,}\footnote{\url{https://github.com/SunriserFuture/EducationQ}}.
\end{abstract}

\section{Introduction}

Large Language Models (LLMs) are revolutionizing various domains, sparking significant interest in their potential to transform education through personalized learning and automated feedback \cite{Memarian2023}. The evolution of LLMs-as-Teachers in applications shifts from simple question-answering to sophisticated teaching capabilities increasingly (Figure~\ref{fig:evolution}). While recent research has explored their applications in specific teaching tasks—including question generation \cite{Olney2023, Shridhar2022}, automated assessment \cite{Nye2023, Patil2024}, feedback provision \cite{Cohn2024}, and teaching support through dialogue \cite{Zha2024, Liu2024}—current benchmarks predominantly assess isolated capabilities like knowledge acquisition, reasoning, and task completion. This narrow focus fails to evaluate core teaching functions essential for effective education: guiding learning processes, facilitating knowledge construction, organizing educational activities, providing personalized feedback, and scaffolding skill development \cite{Palincsar1998, HmeloSilver2006, Mercer2007, Wood1976}.

Existing LLM evaluation approaches - whether through closed-ended questions, open-ended responses, or multi-turn dialogues - present fundamental limitations in assessing teaching capabilities. Current benchmarks predominantly rely on closed-ended assessments, which enable efficient automation but fail to capture the complexity and teacher agency in educational interactions. While open-ended evaluation could better reflect teaching dynamics, it faces significant challenges in scalability and consistency due to reliance on human judgment. Multi-turn dialogue frameworks, despite better capturing interactive complexity, lack specific mechanisms for eliciting and evaluating teaching effectiveness. These limitations particularly impact teaching evaluation: methodologies neither capture teachers’ active role in questioning, assessment, and real-time adaptation, nor provide scalable solutions for evaluating teaching quality.

To address these challenges, we propose EducationQ, a novel multi-agent dialogue framework that incorporates formative assessment into the evaluation of LLMs' teaching capabilities. Formative assessment—a continuous process of assessing learner progress, identifying gaps, and adjusting teaching strategies \cite{Wiliam2011}—is essential for personalized instruction \cite{miao2021ai}. It bridges the gap between students' current abilities and potential, enhances learning outcomes, and promotes educational equity through AI \cite{Pardo2019,RuizPrimo2007,USEducation2023,Allal2000}. In real-world classrooms, these principles are commonly manifested as informal formative assessments (IFAs) during instructional dialogues, where teachers pose questions, assess student understanding, and provide timely feedback and guidance \cite{SezenBarrie2017, Guskey2005}. 

The EducationQ framework models these interactions through a triadic system of teacher, student, and evaluator agents, simulating cyclical teacher-student interaction. This design captures teachers' agency in employing diverse strategies and navigating complex educational contexts while enabling automated evaluation of dialogue quality. To support this framework, we curated a robust dataset of 1,498 questions from established benchmarks GPQA and MMLU-Pro, spanning diverse disciplines and difficulty levels.
We employed a mixed-methods approach to comprehensively evaluate LLMs' teaching capabilities. 
This approach quantifies teaching effectiveness from an outcome-aligned perspective \cite{Gitomer2007} by measuring student learning gains, while analyzing pedagogical strategies using an automated qualitative evaluator agent with strong alignment (78\% agreement) with human education experts.

Our analysis yielded several key findings. Quantitatively, we observed that superior performance in general knowledge benchmarks does not predict teaching effectiveness, with some smaller open-source models outperforming larger commercial ones. And qualitative analysis of teaching dialogues highlights distinct pedagogical strategies contributing to these outcomes.
Our findings reveal model-specific teaching strengths. Llama 3.1 70B Instruct achieved balanced and superior teaching performance through sophisticated questioning strategies, achieving 11.01\% improvement across all evaluation questions and up to 24\% in individual subjects. Gemini 1.5 Pro 002 achieved 7.48\% improvement by providing targeted instructional feedback. OpenAI o1-mini excelled in reasoning-intensive subjects, while Llama 3.1 70B Instruct dominated knowledge-intensive disciplines.

This work advances the field of AI in education through the major contributions:
\begin{itemize}
    \vspace{-0.3cm}
    \itemsep -0.3em 
    \item A theoretical framework integrating formative assessment and Vygotsky's \citeyearpar{Vygotsky1978} learning theory to evaluate educational LLMs.
    \item A multi-agent dialogue methodology for simulating and analyzing teaching interactions.
    \item A high-quality educational dataset comprising standardized tests and re-annotated teacher-student dialogues with pre/post-test results (14,980 five-round interactions).
    \item Vast empirical evaluations demonstrating significant student learning gains (up to 12.63\% improvement on the GPQA Diamond test set).
\end{itemize}

\section{Related Work}
\subsection{LLM Evaluation}

Task-oriented performance benchmarks like MMLU \cite{Hendrycks2020}, MMLU-Pro \cite{Wang2024}, and GPQA \cite{Rein2023} employ closed-ended questions to evaluate domain knowledge and reasoning abilities. MATH \cite{Hendrycks2021} examines mathematical reasoning, while HumanEval \cite{Chen2021} tests programming capabilities. MathChat\cite{Liang2024} evaluates LLMs on multi-turn mathematical interactions, revealing that models struggle with sustained reasoning across multiple dialogue turns.

Instruction following benchmarks such as IFEval \cite{Zhou2023}, FLAN \cite{Wei2022}, Self-Instruct \cite{Wang2023}, and NaturalInstructions \cite{Wang2022} assess LLMs' ability to comprehend and execute directives through open-ended responses. 

Human preference alignment benchmarks like MT-Bench and Chatbot Arena \cite{Zheng2023} evaluate interaction quality through human judgment, they prioritize general user satisfaction over educational outcomes. 

\subsection{LLM-Enhanced Benchmark Development}
Recent research has increasingly incorporated LLMs as agents in benchmark datasets, tasks, and analysis. For instance, MMLU-Pro employs GPT-4-Turbo to expand distractor options, enhancing test stability \cite{Wang2024}. Benchmarks Self-Evolving \cite{WangS2024} utilizes LLMs to extend existing benchmark sets, reducing data contamination while increasing stability and granularity. Dr.Academy \cite{Chen2024} leverages GPT-4 to evaluate generated content's consistency, relevance, coverage, and representativeness.

LLMs' human-like behavior has led to their use in simulating human judgment, test-taking, and feedback provision. Zheng et al. \citeyearpar{Zheng2023} demonstrated how human-aligned GPT-4 could replace human judges in MTBench, reducing crowdsourcing costs while maintaining evaluation quality.

\subsection{LLM-Based Student Modeling}
Recent advancements have demonstrated the potential of LLMs to simulate nuanced human cognitive processes and behaviors \cite{Park2023GenerativeAgents}, including emulating reasoning pathways \cite{Wei2022, Kojima2022}, and mimicking human-like reflective thinking \cite{weng-etal-2023-large}. This burgeoning ability to model human-like thought and action has naturally led to applying LLMs to simulate student behavior and interactions within educational contexts.
Xu \& Zhang \citeyearpar{Xu2023} investigated the feasibility of using generative students to test educational materials, while Markel et al. \citeyearpar{Markel2023} employed LLMs to simulate student dialogues for teacher training. Further, Lu \& Wang \citeyearpar{Lu2024} found that profile-based generative students can closely mirror human student performance in MCQ responses.
Jin et al. \citeyearpar{Jin2024} proposed TeachTune, a framework generating pedagogical agent dialogues with diverse simulated student profiles for human evaluation, complementing our automated fixed-student-model assessment approach.

\section{Dataset}
We constructed our evaluation datasets, as summarized in Table~\ref{tab:dataset-stats}, by systematically curating questions from two well-established benchmarks: GPQA (n=448), featuring domain expert-authored questions, and MMLU-Pro (n=12,032), containing reasoning-intensive questions with enhanced robustness through 10-option design across 14 categories. These datasets span undergraduate to PhD-level content, providing a rigorous foundation for evaluating teaching capabilities. The resulting dataset benefits from the well-established credibility of source benchmarks, proven stability of their evaluation protocols, and comprehensive representation across multiple teaching dimensions.

\begin{table}[h]
    \centering
    \renewcommand{\arraystretch}{1.3}
    \small
    \begin{tabular}{>{\arraybackslash}m{1.65cm}>{\arraybackslash}m{0.8cm}}
        \hline
        \textbf{Data Source} & \textbf{Count}   \\
        \hline
        GPQA             & \normalsize{448}              \\
        MMLU-Pro         & \normalsize{12,032}           \\
        & \\
        \hline
    \end{tabular}
    \begin{tabular}{>{\arraybackslash}m{2.8cm}>{\arraybackslash}m{0.6cm}}
        \hline
        \textbf{Extracted Dataset} & \textbf{Count} \\
        \hline
        GPQA D\tiny{IAMOND}             & \normalsize{198}            \\
        MMLU-Pro S\tiny{TRATIFIED}      & \normalsize{1,300}          \\
        {\bfseries Total}        & \normalsize{1,498}          \\
        \hline
    \end{tabular}
    \vspace{-0.2cm}
    \caption{Dataset construction and distribution statistics.}
    \label{tab:dataset-stats}
    \vspace{-0.2cm}
\end{table}

\begin{figure}[ht]
    \vspace{-0.3cm}
    \includegraphics[width=0.495\linewidth]{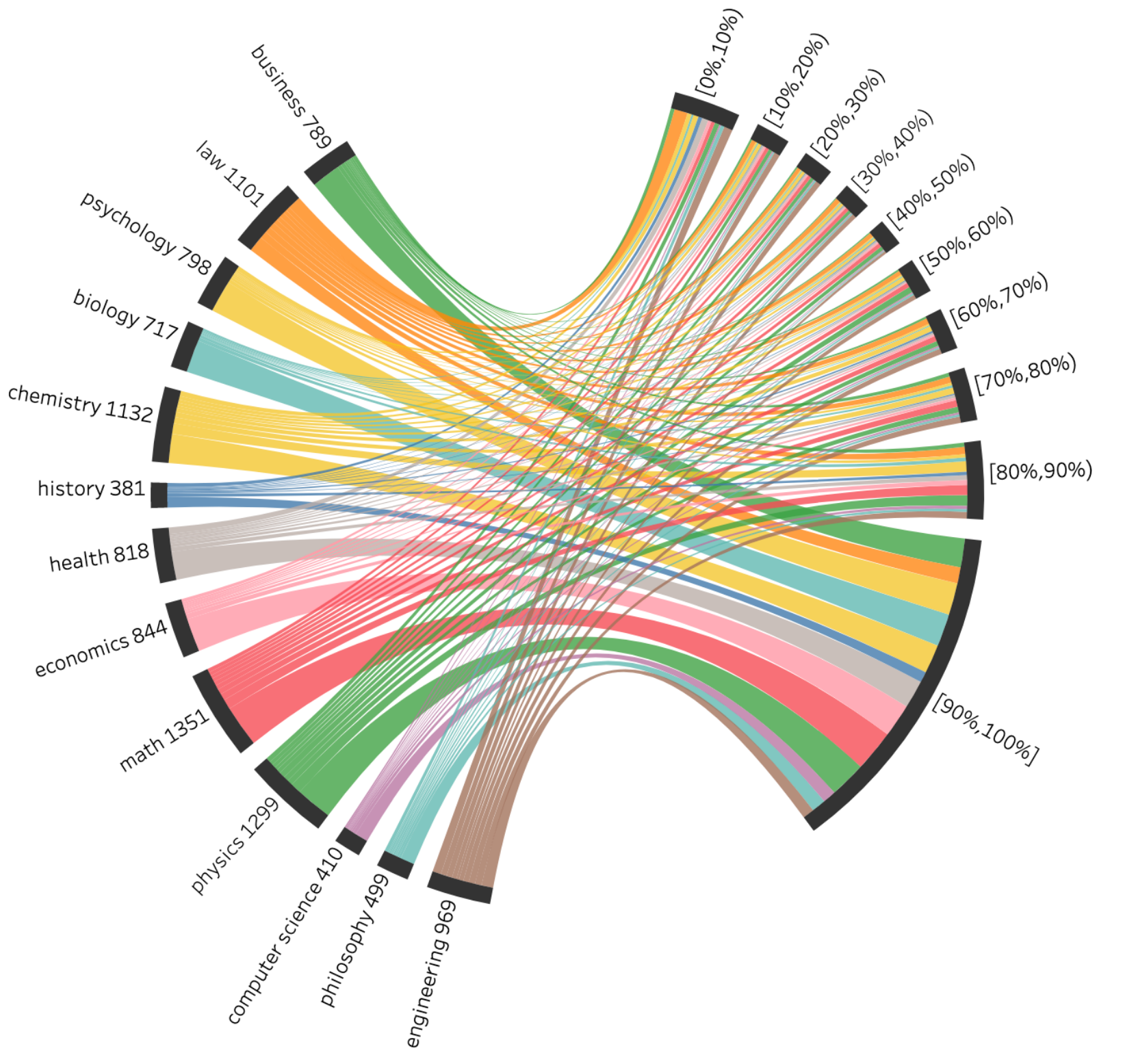} \hfill
    \includegraphics[width=0.495\linewidth]{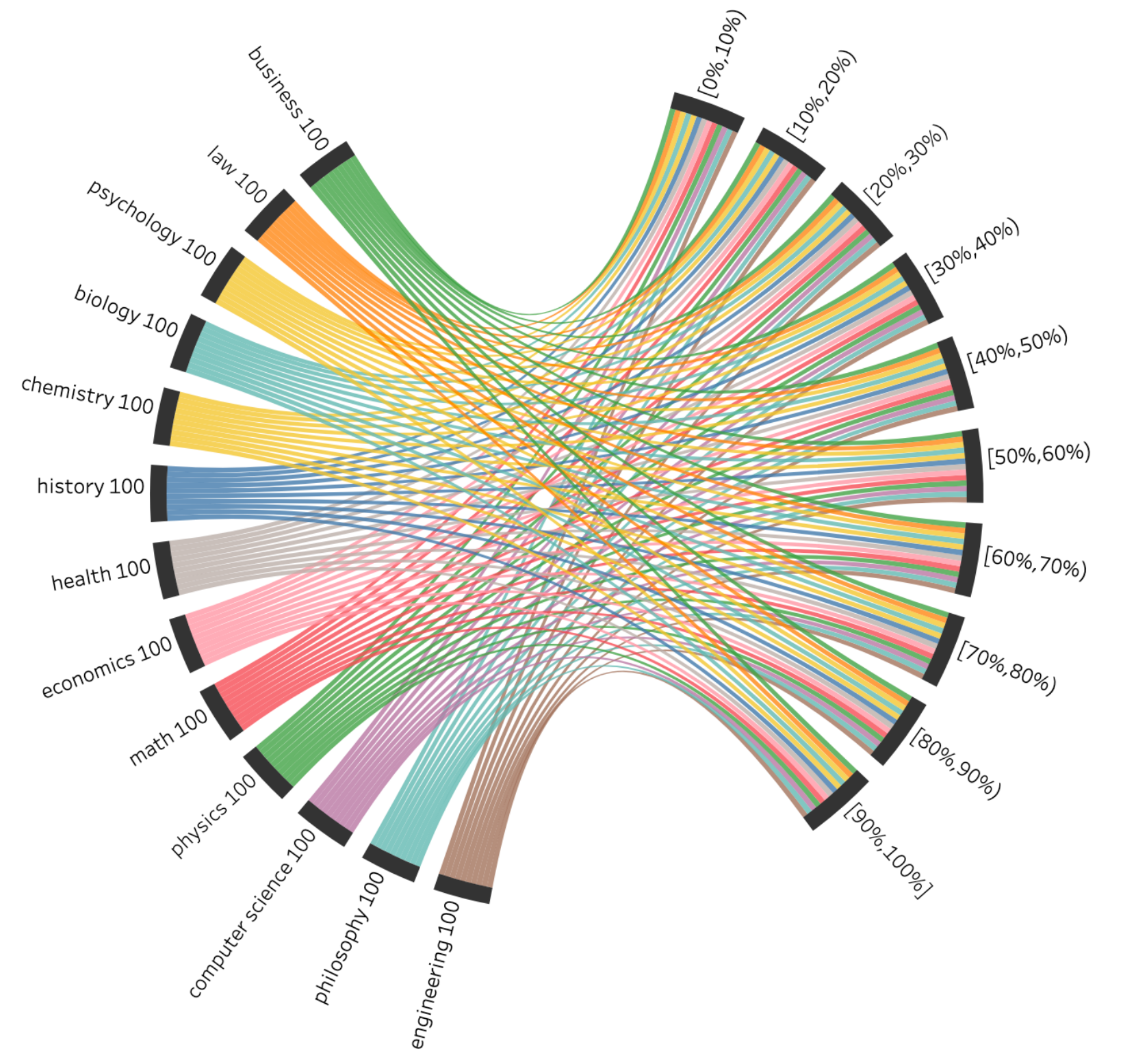}
    \vspace{-0.7cm}
    \caption {Dataset distribution across 13 academic disciplines and 10 difficulty levels. Left: original MMLU-Pro shows uneven distribution across subjects and difficulty levels; Right: MMLU-Pro Stratified with balanced sampling approach ensuring equal representation.}
    \label{fig:mmlu-pro-stratified-dist}
    \vspace{-0.2cm}
\end{figure}

To optimize both assessment quality and efficiency, we focused on two carefully selected subsets: (1) GPQA Diamond (n=198), an expert-validated subset of GPQA with empirically verified difficulty (demonstrated by < 33\% correct response rate among non-experts), and (2) our newly constructed MMLU-Pro Stratified (n=1,300).We developed MMLU-Pro Stratified through systematic sampling based on performance analysis of the top 10 models from published evaluation results\footnote{https://github.com/TIGER-AI-Lab/MMLU-Pro.} (accessed September 2024). As visualized in Figure~\ref{fig:mmlu-pro-stratified-dist}, we calculated mean accuracy rates across all valid responses for each question, excluding null or malformed outputs, to assign difficulty ratings. After removing the "other" category to ensure disciplinary clarity, we stratified the remaining questions into 10 difficulty levels using 10\% intervals and sampled the first 10 questions from each subject-difficulty combination.

The 1,498-question dataset attained a 47.73\% baseline accuracy with Llama 3.1 70B Instruct as the student agent, providing a reference for teaching effectiveness. The distribution simulates diverse educational scenarios, with balanced representation across difficulty tiers and disciplines ensuring comprehensive analytical coverage.

\section{EducationQ Multi-Agent Framework}
\begin{figure*}[b!]
\vspace{-0.4cm}
    \centering
    \includegraphics[width=0.91\textwidth]{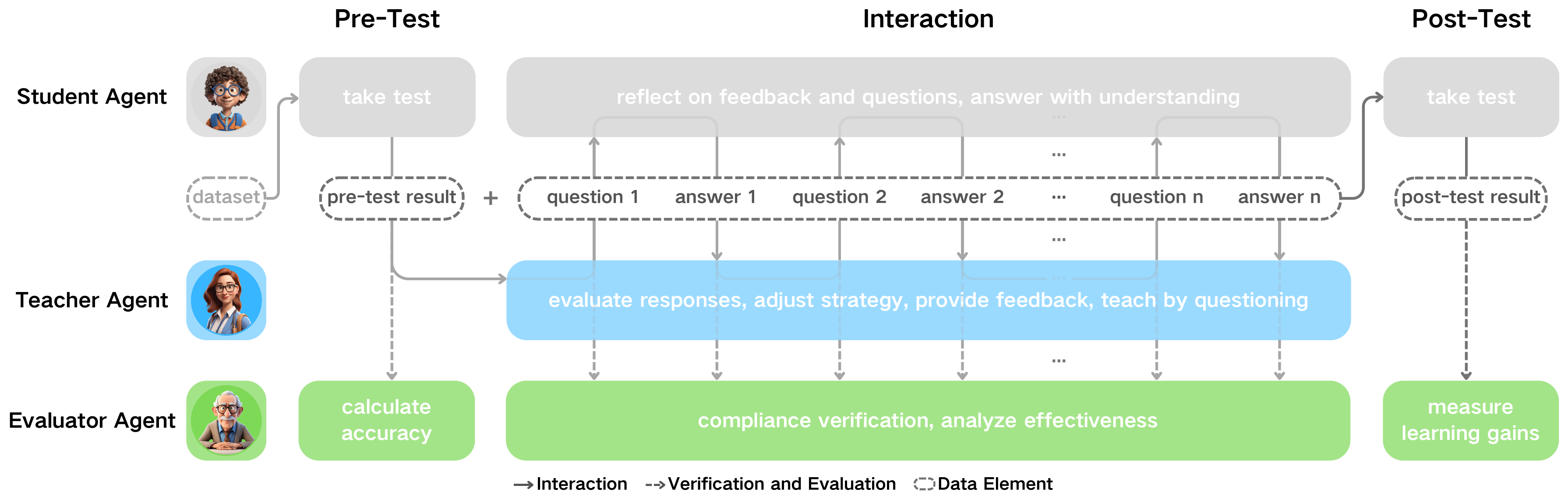}
    \vspace{-0.3cm}
    \caption{The formative assessment interaction flow in the EducationQ framework, detailing the multi-agent multi-turn dialogue implementation shown in Figure 1(c).}
    \vspace{-0.1cm}
    \label{fig:interaction-flow}
\end{figure*}

Our methodology employs three distinct agents: the teacher agent under evaluation, the student agent participating in standardized tests and IFA dialogues, and the evaluator agent providing analysis, as illustrated in Figure~\ref{fig:interaction-flow}.

\begin{table*}[b!]
\centering
\renewcommand{\arraystretch}{1.2}
\scriptsize
\setlength{\tabcolsep}{3.6pt}
\begin{tabular}{l|*{2}{>{\centering\arraybackslash}p{0.48cm}>{\centering\arraybackslash}p{0.48cm}>{\centering\arraybackslash}p{0.48cm}|>{\centering\arraybackslash}p{0.48cm}>{\centering\arraybackslash}p{0.48cm}>{\centering\arraybackslash}p{0.48cm}|}>{\centering\arraybackslash}p{0.48cm}>{\centering\arraybackslash}p{0.48cm}>{\centering\arraybackslash}p{0.48cm}|>{\centering\arraybackslash}p{0.48cm}>{\centering\arraybackslash}p{0.48cm}>{\centering\arraybackslash}p{0.48cm}}
   \hline
   \multirow{3}{*}{\diagbox[width=2.6cm,height=1cm]{\textbf{Student}}{\textbf{Teacher}}} & 
   \multicolumn{6}{c|}{\parbox{3cm}{\centering\textbf{Llama 3.1 \tiny{70B Instruct}}}} & 
   \multicolumn{6}{c|}{\parbox{3cm}{\centering\textbf{Qwen 2.5 \tiny{72B Instruct}}}} & 
   \multicolumn{6}{c}{\parbox{3cm}{\centering\textbf{Mistral \tiny{Nemo}}}} \\
   \cline{2-19}
   & \multicolumn{3}{c|}{Accuracy (\%)} & \multicolumn{3}{c|}{Metrics} & 
     \multicolumn{3}{c|}{Accuracy (\%)} & \multicolumn{3}{c|}{Metrics} &
     \multicolumn{3}{c|}{Accuracy (\%)} & \multicolumn{3}{c}{Metrics} \\
   \cline{2-19}
   & Pre & Post & $\Delta$ & PNIR & CSS & UIC & Pre & Post & $\Delta$ & PNIR & CSS & UIC & Pre & Post & $\Delta$ & PNIR & CSS & UIC \\
   \hline
   \textbf{Llama 3.1 \tiny{70B Instruct}} & 46.97 & 59.60 & 12.63 & 0.26 & 0.26 & 22 & 46.97 & 55.05 & 8.08 & 0.27 & 0.26 & 8 & 46.97 & 51.52 & 4.55 & 0.47 & 0.24 & 5 \\
   \textbf{Qwen 2.5 \tiny{72B Instruct}} & 45.45 & 54.04 & 8.59 & 0.06 & 0.22 & 13 & 45.45 & 50.00 & 4.55 & 0.18 & 0.24 & 4 & 45.45 & 47.98 & 2.53 & 0.17 & 0.25 & 2 \\
   \textbf{Mistral \tiny{Nemo}} & 35.35 & 42.42 & 7.07 & 0.42 & 0.18 & 17 & 35.35 & 37.88 & 2.53 & 0.72 & 0.19 & 13 & 35.35 & 35.35 & 0.00 & 1.00 & - & 8 \\
   \hline
\end{tabular}
\begin{tablenotes}\scriptsize
   \item Dataset: GPQA Diamond, Pre: Pre-test accuracy, Post: Post-test accuracy, $\Delta$: Absolute learning gain, PNIR: Positive-Negative Impact Ratio (lower is better), CSS: Cross-subject Stability (lower is better), UIC: Unique Improvement Count.
\end{tablenotes}
\vspace{-0.2cm}
\caption{Student Agent Ablation Study Based on GPQA Diamond.}
\vspace{-0.2cm}
\label{tab:student-ablation-study}
\end{table*}

\subsection{Student Agent}
The student agent is initiated and prompted (see Appendix~\ref{app:student_prompt}) to focus on specific subjects, analyze problems, and express thoughts and uncertainties, mimicking authentic student behavior. We implement soft token limits rather than hard cutoffs to maintain natural response patterns. Llama 3.1 70B Instruct (GPQA Diamond 46.97\%) serves as our student agent due to its open-source availability for reproducibility, strong instruction-following capabilities (86.96 IFEval), and balanced performance-cost ratio at 70B parameters.

Ablation studies, as showed in Table~\ref{tab:student-ablation-study} using Qwen 2.5 72B Instruct (IFEval 86.38; GPQA Diamond 45.45\%) and Mistral Nemo 12b (IFEval 62.03; GPQA Diamond 35.35\%) as alternative student models showed negligible impact on experimental rankings, suggesting our methodology effectively isolates teacher model performance differences independent of student model selection.

\subsection{Teacher Agent}
Teacher agents are initiated and prompted (see Appendix~\ref{app:teacher_prompt}) to conduct dynamic assessment of student thinking processes and dialogue performance, employing probing questions to gauge understanding and promote thinking, providing feedback, and offering necessary corrections\cite{SezenBarrie2017}.

To prevent direct answer disclosure, we implement strict data flow controls that technically prevent teacher agents from accessing any of the question options. This enforced constraint, combined with explicit prompt instructions, ensures teachers must guide learning without revealing answers.

\subsection{Evaluator Agent}
The evaluator agent is prompted (see Appendix~\ref{app:evaluator_configuration}) as an education assessment expert well-versed in pedagogical theory and practice. It validates the dialogues and evaluates teaching according to specified dimensions and compares teacher performances to determine superior approaches.

The qualitative analysis framework comprises 17 distinct scoring dimensions, including teacher-focused metrics (questioning, assessment, feedback) and student-impact measures (metacognitive reflection, knowledge dimension, etc.) \cite{Krathwohl2002,Looney2011,Wilen1987,Wass2014}. Given the exploratory nature of this component, we did not address the dimensional overlap.

\subsection{Interaction Protocol}
Our teaching interaction design simulates informal formative assessment (IFA) scenarios in classroom settings. In these contexts, the boundaries between curriculum, instruction, and assessment become fluid \cite{Duschl1997}, with teachers enhancing students' mastery of learning objectives through continuous dialogue, ultimately leading to improved summative assessment performance \cite{RuizPrimo2007}.

In our framework, pre-test and post-test correspond to standardized summative assessments, while the multi-turn interactions represent classroom IFAs. The difference in accuracy between these assessments, termed Absolute Learning Gain (ALG, Equation~\ref{eq:alg}), reflects student performance changes before and after teacher dialogue \cite{McGrath2015}, providing a reliable measure of overall teaching effectiveness.

\subsection{Pre-Test}
The pre-test establishes the student agent's initial knowledge baseline while providing teachers with preliminary insights through chain-of-thought reasoning patterns. To ensure broader applicability and stability, we conducted pre-test evaluation following the official MMLU-Pro and GPQA Diamond benchmark protocols and parameters.

\subsection{Interaction}
Dialogues proceed question by question, with teachers receiving message-format access to question content, student responses, and correctness judgments before initiating the first interaction round. Each teacher-student exchange constitutes one round, with five rounds per question.

\subsection{Post-Test}
To maintain compatibility with existing benchmarks, we employed MMLU-Pro and GPQA evaluation protocols rather than student agent assessment. The post-test incorporated pre-test reasoning records and subsequent teacher-student dialogue content via message format while maintaining consistent parameter settings.

\subsection{Content Boundary Design}
As mentioned in the Teacher Agent section, to prevent direct answer disclosure, we implemented the following technical constraints: (a) Teacher agents cannot access any of the answer options, relying solely on student reasoning patterns and correctness judgments for guidance. (b) Students cannot access pre-test correctness judgments during dialogue, learning exclusively through teacher interaction. (c) Students retain access to complete question content including options, enabling learning through experience association.

\subsection{Interaction Parameters and Constraints}
Ablation studies identified optimal parameters of 150 tokens per turn across five rounds, balancing effectiveness and efficiency. Increasing token limits to 250 yielded no significant learning gains, while reducing teacher dialogue to 70-100 tokens degraded teaching performance. Doubling rounds to 10 (with halved student token limits) increased computational costs without surpassing the effectiveness of the 5-round, 150-token configuration. In final experiments, teacher responses averaged 73.6 tokens, with student responses averaging 260 tokens.

\subsection{Data Quality Verification}
Two automated retry mechanisms ensure data integrity: (1) Empty Response Detection: Triggers on zero token count, indicating model output failure. (2) Anomalous Output Detection: Activates when token counts significantly exceed normal ranges (>80\% of 1024 tokens for dialogue or >80\% of 2048 tokens for test answers). 

These mechanisms automatically retry with a five-attempt limit per question. Normal response token counts averaged: 73.6 for teacher dialogue, 260 for student responses, and 425 for test answers. All retry-triggering cases underwent manual review for root cause analysis and validation, ensuring interaction data reliability.

\section{Evaluation Metrics}
\label{sec:evaluation_metrics}
We developed a comprehensive evaluation framework considering both quantitative performance and teaching stability:

\begin{enumerate}
    \itemsep -0.3em 
    \item Absolute Learning Gain (ALG): Measures the direct improvement in student performance:
        \vspace{-0.2cm}
        \begin{equation}
            ALG = ACC_{post} - ACC_{pre}
            \label{eq:alg}
        \vspace{-0.2cm}
        \end{equation}
        where $ACC_{post}$ and $ACC_{pre}$ represent the accuracy scores in post-test and pre-test, respectively. This metric reflects the overall teaching effectiveness and enables direct comparison with conventional benchmarking methods.

    \item Positive-Negative Impact Ratio (PNIR): Evaluates the consistency of teaching effectiveness:
        \vspace{-0.3cm}
        \begin{equation}
           PNIR = \frac{N_{neg}}{N_{pos}}
           \label{eq:pnir}
        \vspace{-0.2cm}
        \end{equation}
        where $N_{neg}$ and $N_{pos}$ represent the number of negative and positive teaching impact cases, respectively. Lower PNIR indicates more stable teaching performance.

    \item Cross-subject Stability (CSS): Measures the standard deviation of learning gains across subjects:
        \vspace{-0.5cm}
        \begin{equation}
            CSS = \sigma(SLGPD)
            \label{eq:css}
        \vspace{-0.3cm}
        \end{equation}
        where $\sigma$ denotes the standard deviation and $SLGPD$ represents Subject-wise Learning Gains Percentage Distribution. A lower $CSS$ value indicates more consistent cross-subject teaching capability.

    \item Unique Improvement Count (UIC): Identifies questions where only one specific teacher model achieved improvement:
        \vspace{-0.3cm}
        \begin{equation}
            UIC = Count(QUI)
            \label{eq:uic}
        \vspace{-0.3cm}
        \end{equation}
        where $QUI$ denotes the set of Questions with Unique Improvement, representing cases where only a single teacher model demonstrated enhanced performance. This metric helps identify specialized teaching capabilities of different models.
    
\end{enumerate}

\section{Experimental Setup}
We evaluated both open-source and commercial LLMs across different organizations and scales, selecting models with varying performance levels on MMLU and GPQA benchmarks to ensure comprehensive coverage.

All experiments were conducted through APIs of online providers, with provider selection based on documented performance metrics. Detailed specifications are provided in Appendix~\ref{app:models}.

Our experiments generated 19,474 valid dialogue sequences across 1,498 questions, along with 5,032 qualitative analyses from the evaluations of 296 dialogues across 17 educational dimensions. Cases are showed in Appendix~\ref{app:case_studies}.

\section{Results}

This section analyzes the evaluated LLMs' teaching performance using the EducationQ framework with a mixed-methods approach combining student-outcome-based quantitative metrics with evaluator-based qualitative analysis, revealing insights into overall effectiveness, robustness, subject-specific strengths, and underlying pedagogical strategies.

\subsection{Overall Quantitative Performance}
As presented in Table~\ref{tab:results}, we evaluated 14 LLMs' teaching effectiveness (see Appendix~\ref{app:models} for specifications) using the EducationQ framework, measured by quantitative metrics (see Section~\ref{sec:evaluation_metrics}).

Llama 3.1 70B Instruct demonstrated superior teaching capability, achieving an average ALG of 11.01\% across the 1,498 questions spanning 13 disciplines. Gemini 1.5 Pro 002 followed closely with a 7.48\% improvement. Notably, these results challenge the assumption that larger model scale directly translates to teaching effectiveness, suggesting that effective teaching strategies, rather than just model size, are critical. 

\subsection{Framework Robustness and Stability}

\begin{table}[b!]
   \centering
   \vspace{-0.4cm}
   \scriptsize
   \setlength{\tabcolsep}{3pt}
   \renewcommand{\arraystretch}{1.2}
   \begin{threeparttable}
       \begin{tabular}{@{}
           >{\raggedright\arraybackslash}p{2.02cm} | 
           >{\centering\arraybackslash}p{0.8cm} | 
           >{\centering\arraybackslash}p{0.7cm} 
           >{\centering\arraybackslash}p{0.8cm} | 
           >{\centering\arraybackslash}p{0.55cm} 
           >{\centering\arraybackslash}p{0.65cm} | 
           >{\centering\arraybackslash}p{0.9cm}@{}} 
           \hline
           \textbf{Teacher Model} &
           $ACC_{\text{pre}}$ &
           \multicolumn{2}{c|}{$ACC_{\text{post}}$} &
           \multicolumn{2}{c|}{$ALG$} &
           \multirow{2}{*}{\normalsize{$\sigma^2$}} \\ 
           \cline{2-6}
           & \textbf{R1/R2} & \textbf{R1} & \textbf{R2} & \textbf{R1} & \textbf{R2} & \\
           \hline
           \tiny{Llama 3.1 70B Instruct}    & \tiny{43.08036} & \tiny{50.00000} & \tiny{50.00000} & \tiny{6.91964} & \tiny{6.91964} & \tiny{0.00000} \\
           \tiny{Llama 3.1 405B Instruct}   & \tiny{43.08036} & \tiny{48.66071} & \tiny{48.88393} & \tiny{5.58036} & \tiny{5.80357} & \tiny{0.01246} \\
           \tiny{Claude 3.5 Sonnet}         & \tiny{43.08036} & \tiny{47.76786} & \tiny{47.99107} & \tiny{4.68750} & \tiny{4.91071} & \tiny{0.01246} \\
           \hline
           \multicolumn{6}{@{}r|}{\textit{Mean Variance}} & \textit{\tiny{0.00832}} \\
           \hline
       \end{tabular}
       \begin{tablenotes}[flushleft]\tiny
            \item Note: R1/R2: First/Second run. {$ACC_{\text{pre}}$/$ACC_{\text{post}}$/$ALG$} (in \%): defined in Equation\ref{eq:alg}. $\sigma^2$: Variance between R1 and R2 ALG results. Dataset: GPQA-main (N=448). Student Model: Llama 3.1 70B Instruct.
       \end{tablenotes}
   \end{threeparttable}
   \vspace{-0.2cm}
   \caption{Stability Study of the Multi-agent Framework.}
   \label{tab:stability_analysis}
\end{table}

Our evaluation framework demonstrates strong robustness and reliability including cross-dataset consistency and test-retest reliability. 

First, we observe high cross-dataset consistency (r=0.871, p<0.001) in model rankings between GPQA Diamond and MMLU-Pro Stratified, indicating the framework’s stable evaluation capability regardless of question source.

Second, our framework consistently identifies subject-specific teaching strengths across datasets. Teacher models rankings show strong correlations between GPQA Diamond and MMLU-Pro Stratified in Physics (r=0.904) and Chemistry (r=0.917), with moderate correlation in Biology (r=0.625, limited by GPQA Diamond’s smaller sample of 19 Biology questions). This cross-dataset consistency confirms the framework’s reliability in evaluating subject-specific teaching capabilities.

To further validate framework stability, as shown in Table~\ref{tab:stability_analysis}, we conducted repeated evaluations using GPQA-main (N=448) with three representative models under identical conditions. The low mean variance of 0.00832 in ALG across runs indicates high measurement consistency for the framework.

\begin{table*}[htbp]
   \centering
   \begin{threeparttable}
   \caption{Teaching Performance Comparison of Large Language Models.}
   \vspace{-0.5cm}
   \footnotesize
   \renewcommand{\arraystretch}{1.1}
   \setlength{\tabcolsep}{4pt}
   \begin{tabular}{l|*{3}{>{\centering\arraybackslash}m{0.73cm}>{\centering\arraybackslash}m{0.73cm}>{\centering\arraybackslash}m{0.73cm}|}>{\centering\arraybackslash}m{0.73cm}>{\centering\arraybackslash}m{0.73cm}>{\centering\arraybackslash}m{0.73cm}}
       \hline
       \multirow{3}{*}{\textbf{Model}} & \multicolumn{3}{c|}{\textbf{\scriptsize{GPQA D}\tiny{IAMOND}}} & \multicolumn{3}{c|}{\textbf{\scriptsize{MMLU-Pro S}\tiny{TRATIFIED}}} & \multicolumn{3}{c|}{\textbf{Overall}} & \multicolumn{3}{c}{\textbf{Additional}} \\
       & \multicolumn{3}{c|}{\textbf{Accuracy (\%)}} & \multicolumn{3}{c|}{\textbf{Accuracy (\%)}} & \multicolumn{3}{c|}{\textbf{Accuracy (\%)}} & \multicolumn{3}{c}{\textbf{Metrics}} \\
       \cline{2-13}
       & Pre & Post & \textbf{$\Delta$} & Pre & Post & \textbf{$\Delta$} & Pre & Post & \textbf{$\Delta$} & CSS & PNIR & UIC \\
       \hline
       Llama 3.1 70B Instruct& 46.97 & \textbf{59.60} & \textbf{12.63} & 47.85 & \textbf{58.62} & \textbf{10.77} & 47.73 & \textbf{58.74} & \textbf{11.01} & \underline{0.041} & \textbf{0.18} & \textbf{37} \\
       Gemini 1.5 Pro 002& 46.97 & 54.55 & 7.58 & 47.85 & \underline{55.31} & \underline{7.46} & 47.73 & \underline{55.21} & \underline{7.48} & \textbf{0.030} & 0.40 & \textbf{37} \\
       Llama 3.1 405B Instruct& 46.97 & \textsl{55.05} & \textsl{8.08} & 47.85 & \textsl{53.69} & \textsl{5.85} & 47.73 & \textsl{53.87} & \textsl{6.14} & \textsl{0.045} & \textsl{0.28} & 9 \\
       OpenAI o1-mini& 46.97 & \underline{56.57} & \underline{9.60} & 47.85 & 53.12 & 5.27 & 47.73 & 53.57 & 5.84 & 0.051 & \underline{0.25} & 7 \\
       Qwen 2.5 72B Instruct& 46.97 & \textsl{55.05} & \textsl{8.08} & 47.85 & 52.85 & 5.00 & 47.73 & 53.14 & 5.41 & 0.054 & 0.33 & 7 \\
       Llama 3.1 8B Instruct& 46.97 & 52.02 & 5.05 & 47.85 & 52.69 & 4.85 & 47.73 & 52.60 & 4.87 & 0.051 & 0.40 & \underline{13} \\
       Hermes 3 Llama 3.1 70B& 46.97 & 51.52 & 4.55 & 47.85 & 51.92 & 4.08 & 47.73 & 51.87 & 4.14 & 0.051 & 0.39 & 6 \\
       Mistral Nemo& 46.97 & 51.52 & 4.55 & 47.85 & 51.69 & 3.85 & 47.73 & 51.67 & 3.94 & 0.058 & 0.44 & \textsl{12} \\
       Claude 3.5 Sonnet& 46.97 & 52.53 & 5.56 & 47.85 & 51.38 & 3.54 & 47.73 & 51.54 & 3.81 & 0.059 & 0.30 & 5 \\
       WizardLM-2 8x22B& 46.97 & 50.51 & 3.54 & 47.85 & 51.54 & 3.69 & 47.73 & 51.40 & 3.67 & 0.047 & 0.34 & 2 \\
       DeepSeek V2.5& 46.97 & 50.51 & 3.54 & 47.85 & 51.08 & 3.23 & 47.73 & 51.00 & 3.27 & 0.051 & 0.46 & 3 \\
       Command R 08-2024& 46.97 & 49.49 & 2.53 & 47.85 & 50.85 & 3.00 & 47.73 & 50.67 & 2.94 & 0.057 & 0.53 & 7 \\
       GPT-4o-mini& 46.97 & 50.51 & 3.54 & 47.85 & 50.12 & 2.27 & 47.73 & 50.17 & 2.44 & 0.085 & 0.40 & 2 \\
       Phi-3.5-mini Instruct& 46.97 & 48.99 & 2.02 & 47.85 & 48.92 & 1.08 & 47.73 & 48.93 & 1.20 & 0.172 & 0.69 & 4 \\
       \hline
   \end{tabular}
       \begin{tablenotes}\footnotesize
           \item Note: Pre: Pre-Test Accuracy; Post: Post-Test Accuracy; $\Delta$: Absolute Learning Gain; 
           CSS: Cross-subject Stability (lower is better); PNIR: Positive-Negative Impact Ratio (lower is better); UIC: Unique Improvement Count.
           The best results are marked in \textbf{bold}, second best results are \underline{underlined}, and third best results are in \textsl{italics}.
       \end{tablenotes}
   \vspace{-0.4cm}
   \label{tab:results}
   \end{threeparttable}
\end{table*}

\subsection{Subject-Specific Performance}

Beyond overall performance, models exhibited distinct subject specializations (detailed rankings in Appendix~\ref{app:category-delta-results-sorted-ranked}). Llama 3.1 70B Instruct excelled in knowledge-intensive subjects, leading in Psychology (ALG=18\%), Health (ALG=24\%), and law (ALG=11\%). OpenAI o1-mini dominated Physics (ALG=8.6\%) and Math (ALG=9\%), demonstrating strength in logical reasoning and problem-solving. Gemini 1.5 Pro 002 showed particular prowess in applied disciplines like Business (ALG=8\%) and Economics (ALG=9\%), reflecting superior integration of theoretical knowledge with practical applications. Additionally, Hermes 3 Llama 3.1 70B led in Engineering (ALG=10\%), while Qwen 2.5 72B Instruct topped Chemistry in MMLU-Pro Stratified Subset(ALG=11\%).

Cross-subject stability (CSS, Equation~\ref{eq:css}) reveals that Gemini 1.5 Pro 002 (CSS=0.030) and Llama 3.1 70B Instruct (CSS=0.041) provided the most consistent teaching performance across subjects.

\subsection{Performance Across Difficulty Levels}
Analyzing performance across 10 difficulty levels (derived from MMLU-Pro Stratified baseline accuracies, see Figure~\ref{fig:mmlu-pro-stratified-dist}), Llama 3.1 70B Instruct showed the most stable performance across difficulty levels ($\sigma$=0.032), closely followed by Gemini 1.5 Pro 002 ($\sigma$=0.043). Most LLM teachers performed best with relatively simple questions (prior accuracy \~0.8), with these improvements accounting for approximately 20\% of total gains, suggesting strength in reinforcing well-understood concepts.

However, the Llama 3.1 series (70B and 8B models) exhibited a distinctly different pattern, achieving peak performance at medium difficulty levels (prior accuracy \~0.5), accounting for 27\%(70B) and 19\%(8B) of their ALGs. In contrast, their improvement rates at the easiest level (prior accuracy \~0.8) represented only 11\% of their ALGs. This pattern suggests these models might possess an advantage in scaffolding learning for moderately challenging concepts rather than reinforcing already known zone, demonstrating effectiveness in helping students breakthrough current knowledge boundaries. 

\subsection{Teaching Stability Analysis}
Through analysis of the Positive-Negative Impact Ratio (PNIR, Equation~\ref{eq:pnir}, lower is better), we identified significant variations in teaching stability across models. Llama 3.1 70B Instruct demonstrated exceptional stability, generating only 36 negative cases against 200 positive improvements (PNIR = 0.18). While Gemini 1.5 Pro 002 achieved comparable positive cases (188), its higher PNIR of 0.40 indicated greater performance volatility. OpenAI o1-mini and Llama 3.1 405B Instruct maintained moderate stability (PNIR = 0.25 and 0.28 respectively). These findings suggest that high teaching effectiveness and high stability can coexist. 

\subsection{Unique Improvement Analysis}
The Unique Improvement Count (UIC, Equation~\ref{eq:uic}), identifying cases where only one specific teacher model produced a learning gain for a given question, highlights specialized capabilities. Gemini 1.5 Pro 002 and Llama 3.1 70B Instruct particularly excelled, each achieving 37 unique improvements. However, their patterns differed: Llama 3.1 70B Instruct were more balanced across disciplines (standard deviation 0.036, peaking at 14\% in Psychology), while Gemini 1.5 Pro 002 showed stronger subject preferences (standard deviation 0.056, reaching 21\% in Biology). OpenAI o1-mini, despite modest overall performance, secured 3 unique improvements specifically in Engineering, hinting at potentially valuable niche expertise. 

\subsection{Evaluator-Based Qualitative Analysis}

As cases in Appendix~\ref{app:case_1057}, we conducted an evaluator-agent analysis of 148 UIC cases and their paired non-improvement control dialogues (296 teacher-student dialogues total) to explore the potential of LLM-based evaluators for nuanced educational dialogue assessment (validated against human experts in Section~\ref{sec:human_eval}) and to identify teaching strategies correlated with successful outcomes.

We used GPT-4o as our evaluator model, due to the high human-alignment of GPT-4 herd models \cite{Zheng2023,Chen2024}. The evaluator agent assessed each dialogue through three distinct analytical perspectives and dimensions, showed in Appendix~\ref{app:evaluation_dimensions}, using a standardized 1-10 scale. The scores were subsequently used as predictors in statistical models, with the binary improvement status (improvement vs. non-improvement) serving as the target variable.

Initial logistic regression analysis revealed significant predictors of  learning gains: questioning quality for Llama 3.1 70B Instruct (Exp(B)=32.864, p=0.043) and feedback quality for Gemini 1.5 Pro 002 (Exp(B)=5227.342, p=0.019). Random forest analysis (1000 trees, accuracy=0.769, AUC=0.775) further identified Llama 3.1 70B Instruct’s effectiveness correlated strongly with questioning strategies (Mean dropout loss 0.363), while Gemini 1.5 Pro 002’s success primarily stemmed from feedback (Mean dropout loss 0.344). 

\section{Expert Analysis and Human Alignment}
\label{sec:human_eval}

While the EducationQ framework enables automated quantitative evaluation of teaching capabilities based on objective student learning outcomes, we introduce expert reviewers to analyze teaching cases and validate the evaluator-agent-based qualitative methodology, demonstrating 78\% alignment between human judgment and the evaluator agent's verdict of effective teaching behaviors. This validation provides deeper insights into the pedagogical approaches employed by LLMs and confirms the reliability of our mixed-methods framework.

\subsection{Teaching Strategy Analysis: Expert-Annotated Case Studies}

\begin{table}[t]
    \centering
    \footnotesize
    \caption{Cases of Teacher-Student Dialogues}
    \label{tab:case_studies}
    \vspace{-0.1cm}
    \begin{tabularx}{0.99\columnwidth}{>{\raggedright\arraybackslash}X}
        \hline \\
        \textbf{D1: Mathematical Reasoning Examples} (Question 240) \\[4pt]
        \textbf{D1.1: Llama 3.1 70B Intruct} \\[2pt]
        \textbf{T:} "Can you explain why you chose to divide..." \\
        \textbf{S:} [response] \\
        \textbf{T:} "... or could it be an approximation? How might this affect your calculation, and what implications would it have for determining the correct answer?" \\
        \\
        \textbf{D1.2: Llama 3.1 405B Instruct} \\[2pt]
        \textbf{T:} "... how much will Daniel's ..." \\
        \textbf{S:} [response] \\
        \textbf{T:} "Can you calculate..." \\[4pt]
        \hline \\
        
        \textbf{D2: Legal Concept Examples} (Question 961) \\[4pt]
        \textbf{D2.1: Gemini 1.5 Pro 002} \\[2pt]
        \textbf{T:} "While taking out an insurance policy could be part of a larger criminal plan, is it in itself a substantial step directly towards committing the crime of murder? Think about the specific actions needed to carry out a murder and whether purchasing insurance moves the wife closer to completing those actions." \\
        \textbf{S:} [response] \\
        \textbf{T:} "Good, you've identified that purchasing the insurance policy is not a substantial step. Now, let's consider the wife's other actions." \\
        \\
        
        \textbf{D2.2: Claude 3.5 Sonnet} \\[2pt]
        \textbf{T:} "... In the context of attempted murder, what constitutes a 'substantial step' towards committing the crime? Can you provide an example of a substantial step that would be more clearly indicative of an attempt to commit murder than purchasing a life insurance policy?..." \\
        \textbf{S:} [response] \\
        \textbf{T:} "...You've provided a good example of a more substantial step. ..." \\
        \\
        \hline
    \end{tabularx}
    \vspace{-0.3cm}
\end{table}

Human education experts confirmed that learning gains resulted from diverse and sophisticated teaching strategies rather than direct answer disclosure. The experts identified and annotated clear evidence of pedagogical techniques. Our expert-annotated case studies provide detailed evidence of how different LLMs implement distinct teaching strategies and demonstrate varying pedagogical skills and explain why models with superior general capabilities might underperform in educational interactions.

As presented in Table~\ref{tab:case_studies} (detailed in Appendix~\ref{sec:case_240},~\ref{sec:case_961}), Llama 3.1 70B Instruct demonstrated sophisticated teaching techniques highly aligned with established educational theory. These cases showed carefully constructed question sequences and metacognitive support scaffolding student understanding. As illustrated in dialogue D1.1, the model employed guidance across different cognitive levels (Bloom’s Taxonomy) rather than focusing solely on procedural practice. This contrasted sharply with Llama 3.1 405B Instruct’s approach to the same problem (D1.2), which, despite greater general capabilities, emphasized repetitive practice over conceptual understanding. Notably, Llama 3.1 70B Instruct’s progressive questioning through “can you explain why” and “how might this affect” constructed cognitive bridges between students’ current understanding and target concepts, exemplifying excellent application of Zone of Proximal Development theory \cite{Vygotsky1978}.

Gemini 1.5 Pro 002 demonstrated strong adaptive teaching capabilities, characterized by precise diagnostic techniques and targeted, specific feedback. In dialogue D2.1, it successfully identified and addressed student misconceptions about legal concepts, using concept-definition-focused questions to prompt reconceptualization and reinforcing academic concept determination through feedback. This focused approach contrasted with Claude 3.5 Sonnet’s broader methodology and formalized feedback (D2.2), which introduced multiple concepts without adequately addressing core misconceptions and provided feedback based solely on task completion. Gemini 1.5 Pro 002’s rapid diagnosis of conceptual misunderstandings, immediate feedback, and timely strategy adjustments demonstrated excellent formative assessment practice.

These analyses corroborate our quantitative findings while elucidating why larger models may underperform in educational tasks. Through systematic dialogue reviews, we observed that while often demonstrating more domain knowledge, larger models might lack the focused, pedagogically sound interaction strategies consistently exhibited by Llama 3.1 70B Instruct and Gemini 1.5 Pro 002.

\subsection{Teaching Behaviors Analysis: Expert Alignment and Validation}
We conducted a human evaluation study with seven qualified educators and one of the authors to validate our evaluator-agent-based qualitative methodology (see Appendix~\ref{app:comparative_analysis}) to identify effective teaching behaviors. 50 pairs of dialogues (each containing one that produced learning gains and one that did not) were randomly selected from the 148 UIC cases, with teacher identities anonymized.
Comparing the human experts' majority preference for each pair against the evaluator agent's verdict for the same pair, we found that human expert preferences aligned with the evaluator agent's selection in 78\% of cases (39/50). Teachers producing learning gains received significantly higher human ratings (average 7.38/10) than the control group (6.41/10). This sample size represents approximately 40\% of 126-question performance gap between our best and worst-performing teachers, indicating substantial representativeness.
This strong alignment, achieved despite the challenging nature of specialized MMLU-Pro content, confirms that our framework reliably quantifies teaching effectiveness in a manner consistent with human educational expertise. Notably, no reviewer detected any instances of teachers directly revealing answers, validating our Content Boundary Design. Evaluation materials and questionnaire design are available in Appendix~\ref{app:human_eval_questionnaire}.

\section{Conclusion}
\vspace{-0.1cm}
Our comprehensive evaluation of LLMs’ teaching capabilities reveals two critical insights, strengthened by human experts: First, smaller open-source models can outperform larger commercial models through effective pedagogical strategies, challenging conventional assumptions about model scale and teaching effectiveness. Second, successful LLMs-as-Teachers excel through focused, goal-oriented interactions and adaptive teaching methods rather than broader knowledge repositories.

These findings suggest a fudamental rethinking of educational LLM development: prioritizing specialized teaching capabilities over general model scaling.
The significant performance variations also highlight the inadequacy of traditional LLM metrics for predicting teaching effectiveness, underscoring the critical need for specialized, interaction-based evaluation frameworks like EducationQ to guide the development of effective AI in education.

\section*{Limitations}
Our study faces several limitations in evaluation framework, test data, and model selection. Regarding the evaluation framework, our one-on-one IFA scenario cannot fully capture the complexity of teaching roles and capabilities in practice, such as managing classroom dynamics or using student dialogue for concept explanation. Our limitation on dialogue rounds prevented comparison of different LLMs’ teaching efficiency in improving ALG.

In terms of model selection, our teacher model choices did not include newer or older versions within the same series, preventing tracking of teaching capability evolution in LLM development. We also excluded multimodal models and specialized educational private models.

While our test set included advanced topics from graduate to PhD levels across multiple disciplines, we did not evaluate LLMs’ teaching performance with lower-grade content, such as elementary or middle school materials.

Alignment with real-world scenarios represents another major limitation, particularly regarding student modeling and simulation fidelity. Despite basic student ablation studies, we did not employ more sophisticated generative student methods to simulate diverse age groups, cognitive levels, backgrounds, and motivations, thus not fully reflecting the complexity of real teaching situations.

While our primary metric, ALG, objectively measures learning outcomes, the automated qualitative analysis, though validated with 78\% human agreement, captures a predefined set of pedagogical dimensions. Further research could explore even broader qualitative aspects or integrate them more directly into a composite teaching score, if deemed necessary.

\section*{Model Content Limitations}

\begin{table}[b]
    \centering
    \renewcommand{\arraystretch}{1.2}
    \footnotesize
    \vspace{-0.3cm}
    \begin{tabular}{>{\raggedright\arraybackslash}p{0.95\columnwidth}}
        \hline
        \textbf{Question ID:} 5048 \\
        \textbf{Topic:} Political Divergence During Vietnam War \\
        \textbf{Content:} Description of War Impact on Society \\
        \textbf{Model Response:} Consistent NoneType Returns \\
        \hline
    \end{tabular}
    \caption{Question Analysis Example}
\end{table}

During experimentation, we observed protential impacts of content policies on model evaluation. Specifically, OpenAI models (including OpenAI o1-mini and GPT-4o-mini) consistently returned NoneType responses when handling questions about the Vietnam War (Question 5048). This phenomenon, occurring only with specific content-model combinations, likely stems from provider content moderation policies.

This observation highlights a crucial limitation of commercial models in academic evaluation: content moderation policies may create gaps or biases in assessing historically or politically sensitive topics. Such constraints require careful consideration when designing educational evaluation frameworks and academic applications.

\section*{Ethics Statement}

This work focuses on evaluating LLMs' teaching capabilities through automated assessment. While our framework demonstrates potential for educational applications, we acknowledge several ethical considerations:

First, our evaluation framework is designed to assess teaching capabilities rather than replace human teachers. The simulated teaching interactions should be viewed as complementary tools for understanding AI systems rather than substitutes for human-student relationships.

Second, we recognize the limitations of our single-student model approach and the potential bias in educational assessment. Our findings should be interpreted within the context of these constraints, particularly when considering real-world applications.

Our dataset is constructed from publicly available benchmarks (GPQA and MMLU-pro) following their respective terms of use and licensing agreements. We ensure proper attribution and usage of these resources in accordance with their intended research purposes.

Finally, we observed content filtering in some commercial models, highlighting the need for transparent discussion of AI systems' limitations in handling sensitive educational topics.

\bibliography{custom}

\appendix
\section{Agent Configurations and Prompts}
\label{app:prompts}
\subsection{Student Agent Configuration}
\label{app:student_prompt}

The student agent uses a consistent prompting template designed to simulate authentic student learning behavior, including expressing uncertainty and focusing on problem analysis. Below we detail the exact configuration and prompting templates used in our experiments for the student agent.

\subsubsection{Implementation Parameters}
\label{app:student_parameters}
In our experiments, we used the following configuration for the student agent (Meta-Llama-3.1-70B-Instruct):

\begin{itemize}[noitemsep,topsep=0pt]
   \item Temperature: 0.0
   \item Maximum tokens for dialogue responses: 1,024
   \item Maximum tokens for test responses: 2,048 
   \item Recommended token limit for dialogue responses: 150
   \item Recommended token limit for test reasoning responses: 1,024
   \item Token rerun threshold percentage: 80\%
   \item Maximum retries per response: 5
   \item Response Rerun Trigger: Activated if token count exceeds token rerun threshold percentage of the respective maximum tokens and also exceeds the recommended token limit.
\end{itemize}

\subsubsection{Base System Message}
\label{app:student_system_message}
The core system message template is provided initially during the multi-round dialogue interaction:
\begin{lstlisting}[
   basicstyle=\ttfamily\small,
   breaklines=true,
   frame=single,
   numbers=none
]
You are a student focusing on [CATEGORY]. Analyze the question carefully, explain your thought process ([TOKEN_LIMIT] tokens or less), and try to apply the concepts you've learned to solve problems. If you're unsure, express your uncertainty and explain your reasoning.
\end{lstlisting}
Note: The \texttt{[TOKEN\_LIMIT]} placeholder corresponds to the recommended token limit for dialogue responses (150 tokens as default) specified in Appendix~\ref{app:student_parameters}. This message guides the student agent's persona and response style during conversations and avoids endless response.

\subsubsection{Response Format}
For test questions, responses are structured as:
\begin{lstlisting}[
   basicstyle=\ttfamily\small,
   breaklines=true,
   frame=single,
   numbers=none
]
Question: [QUESTION_TEXT]
Options:
A. [OPTION_A]
B. [OPTION_B]
C. [OPTION_C]
D. [OPTION_D]
Let's think step by step.
[REASONING_PROCESS]
The answer is (X)
\end{lstlisting}

For dialogue interactions, The student agent's interaction involves receiving context from previous rounds and generating a new response:
\begin{lstlisting}[
   basicstyle=\ttfamily\small,
   breaklines=true,
   frame=single,
   numbers=none
]
% Example Input Context for generating Student Repsonse in Round n:
[SYSTEM_MESSAGE]

[PRE_TEST_INFO]

% History from Round 1:
Teacher: [TEACHER\_QUESTION\_ROUND\_1]
Student: [STUDENT\_RESPONSE\_ROUND\_1]

% History from Round 2:
Teacher: [TEACHER\_QUESTION\_ROUND\_2]
Student: [STUDENT\_RESPONSE\_ROUND\_2]

% History from other previous rounds:
...

% Current prompt for Round n:
Teacher: [TEACHER\_QUESTION\_ROUND\_n]

% LLM generates round n's response here
\end{lstlisting}
Note: \texttt{[SYSTEM\_MESSAGE]} refers to the template shown in Appendix~\ref{app:student_system_message}. \texttt{[PRE\_TEST\_INFO]} is included in the context if the configuration parameter (\texttt{include\_pretest\_info}) is enabled.

The implementation details can be found in the \texttt{StudentLLM} class, specifically in the \texttt{answer\_question()} and \texttt{take\_test()} methods.

\subsection{Teacher Agent Configuration}
\label{app:teacher_prompt}

The teacher agent employs a structured prompting system to conduct dynamic assessment and provide guided instruction based on the student's pre-test performance and ongoing dialogue. Below we detail the configuration and prompting templates used for the teacher models evaluated in our experiments (See Table~\ref{tab:model-specs} for specific model details).

\subsubsection{Implementation Parameters}
\label{app:teacher_parameters}
Common configuration across all teacher models:

\begin{itemize}[noitemsep,topsep=0pt]
   \item Temperature: 0.0
   \item Maximum tokens per response: 1,024
   \item Recommended token limit for questions: 150
   \item Use Few-Shot Examples: false (as default)
   \item Recommended Education Theory: null (as default)
   \item Token rerun threshold percentage: 80\%
   \item Maximum retries per question: 5
   \item Question Rerun Trigger: Activated if token count exceeds token rerun threshold percentage of the maximum tokens \textit{and} also exceeds the recommended question token limit.
\end{itemize}

\subsubsection{Base System Message}
The core system message template used for all teaching interactions:
\begin{lstlisting}[
   basicstyle=\ttfamily\small,
   breaklines=true,
   frame=single,
   numbers=none
]
You are an expert teacher in [CATEGORY] dedicated to enhancing the student's understanding after analyzing the student's response to a pre-test.

Your task is to ask [NUM_ROUNDS] rounds of relevant, thought-provoking questions to the student. You should ask one new question per round (and if needed, provide necessary corrections or feedback for the student's previous round's answers), each under [TOKEN_LIMIT] tokens, without revealing the correct answers or specific details of the pre-test questions.

Your goal is to prepare the student for the post-test by fostering a deeper and more comprehensive understanding of the subject matter.

Here are the pre-test results of the student: [PRE_TEST_INFO]
\end{lstlisting}
Note: \texttt{[TOKEN\_LIMIT]} corresponds to the recommended question token limit (150 tokens default, see Appendix~\ref{app:teacher_parameters}). \texttt{[PRE\_TEST\_INFO]} is populated using the format described in Appendix~\ref{app:pretest_info}.

\subsubsection{Pre-test Information Format}
\label{app:pretest_info}
The \texttt{[PRE\_TEST\_INFO]} placeholder within the system message is populated with the student's performance on the initial test question, formatted as follows for each pre-test result provided:
\begin{lstlisting}[
   basicstyle=\ttfamily\small,
   breaklines=true,
   frame=single,
   numbers=none
]
Question ID: [ID]
Question: [QUESTION_TEXT]
Student's Reasoning: [REASONING]
Student's Answer: [ANSWER]
Student's Answer is Correct or Not: [CORRECTNESS]
\end{lstlisting}
Note: The \texttt{[CORRECTNESS]} is either "Correct." or "Incorrect." based on the student's selection.

\subsubsection{Interaction Format}
To generate the teaching interaction, the teacher agent receives the system message (including pre-test info) followed by the dialogue history and the final prompt asks for the next question:
\begin{lstlisting}[
   basicstyle=\ttfamily\small,
   breaklines=true,
   frame=single,
   numbers=none
]
% Example Input Context for generating Teacher Interaction in Round n:
[SYSTEM_MESSAGE]

% History from Round 1:
Teacher: [TEACHER\_QUESTION\_ROUND\_1]
Student: [STUDENT\_RESPONSE\_ROUND\_1]
% History from Round 2:
Teacher: [TEACHER\_QUESTION\_ROUND\_2]
Student: [STUDENT\_RESPONSE\_ROUND\_2]

% History from other previous rounds:
...

Teacher: Generate the round [N] question ([TOKEN_LIMIT] tokens or less) to promote better understanding:

% LLM generates round n's teaching here
\end{lstlisting}
Note: \texttt{[TOKEN\_LIMIT]} refers to the recommended question token limit. In the API call, the history turns are mapped to \{\texttt{"role"}: \texttt{"assistant"}, \texttt{"content"}: \texttt{"Teacher: ..."}\} and \{\texttt{"role"}: \texttt{"user"}, \texttt{"content"}: \texttt{"Student: ..."}\} messages, while the final generation prompt is the content of the last \{\texttt{"role"}: \texttt{"user"}\} message.

The implementation details can be found in the \texttt{TeacherLLM} class, specifically in the \texttt{generate\_question()} method.

\subsection{Evaluator Agent Configuration}
\label{app:evaluator_configuration}

The evaluator agent is configured as an expert in educational assessment, providing detailed analysis across multiple dimensions using structured JSON output. The evaluation employs three distinct analytical perspectives, each focusing on different aspects of the teaching-learning process and utilizing specific parts of the interaction data as input. Below we detail the exact configuration, prompts, and evaluation framework used in our experiments.

\subsubsection{Implementation Parameters}
Configuration for the evaluator agent (GPT-4o):
\begin{itemize}[noitemsep,topsep=0pt]
    \item Temperature: 0.0
    \item Maximum tokens per response: 4,096
    \item Response format: Structured JSON schema (\texttt{response\_format=\{ "type": "json\_schema", ... \}})
\end{itemize}

\subsubsection{Base System Message}
\label{app:evaluator_prompt}
The core system message template defines the evaluator's expert role. Specific instructions regarding the dimensions and input data vary depending on the evaluation task (Holistic, Teacher-centric, or Student-centric). A representative template is:
\begin{lstlisting}[
   basicstyle=\ttfamily\small,
   breaklines=true,
   frame=single,
   numbers=none
]
You are an expert in educational assessment with a deep understanding of learning theories and pedagogical practices. Your task is to evaluate the teaching effectiveness based on provided teacher-student interaction data (which may include pre-test results and the dialogue). Carefully analyze the interaction according to the specific dimensions and instructions provided for the current evaluation task. Output your evaluation in the specified structured JSON format, adhering strictly to the provided JSON schema.
\end{lstlisting}
Note: The specific dimensions and required JSON schema vary depending on different aspects of evaluation task (Holistic, Teacher-centric, or Student-centric), as detailed below and implemented in the \texttt{EvaluatorLLM} class.

\subsubsection{Evaluation Dimensions}
\label{app:evaluation_dimensions}
The evaluator assesses effectiveness based on dimensions specific to each analytical perspective:

\paragraph{Holistic Interaction Analysis Dimensons}
\begin{enumerate}[label=(\arabic*), noitemsep,topsep=0pt]
\item Assessment Effectiveness
\item Questioning Effectiveness
\item Feedback Effectiveness
\item Instructional Adaptation Effectiveness
\item Learning Objective Achievement Effectiveness
\end{enumerate}
(Input Data: Full Teacher-Student Dialogue + Pre-test Context.)

This perspective evaluates the overall effectiveness of the teaching interaction based on the complete dialogue, used in \texttt{over\_interaction\_analysis()}.

\paragraph{Teacher-Centric Question Analysis Dimensions}
\begin{enumerate}[label=(\arabic*), noitemsep,topsep=0pt]
    \item Question Relevance
    \item Cognitive Level
    \item Knowledge Dimension
    \item Question Diversity
    \item Scaffolding Progression
    \item Metacognitive Promotion
\end{enumerate}

(Input Data: Only Teacher Questions from the Dialogue.)

This perspective focuses solely on the quality and characteristics of the questions posed by the teacher, used in \texttt{teacher\_questions\_analysis()}.

\paragraph{Student-Centric Response Analysis Dimensions}

\begin{enumerate}[label=(\arabic*), noitemsep,topsep=0pt]
    \item Response Relevance
    \item Cognitive Level Demonstration
    \item Knowledge Dimension Integration
    \item Response Diversity
    \item Elaboration Progression
    \item Metacognitive Reflection
\end{enumerate}

(Input Data: Only Student Answers from the Dialogue.)

This perspective assesses the student's performance and understanding as demonstrated in their responses, serving as an indirect measure of the teacher's impact, used in \texttt{student\_responses\_analysis()}.

\subsubsection{Evaluation Input Format (Example: Holistic Interaction Analysis)}

Input provided to the evaluator for comparative analysis, with teachers anonymized as \texttt{teacher\_a} and \texttt{teacher\_b}:

\begin{lstlisting}[
   basicstyle=\ttfamily\small,
   breaklines=true,
   frame=single,
   numbers=none
]
Question ID: [ID]
Category: [CATEGORY]

Pre-test Result:
[JSON_DUMP_OF_PRE_TEST_RESULT(S)]

<|The Start of teacher_a's Interaction with Student|>
[FORMATTED_INTERACTION_TEACHER_A]
<|The End of teacher_a's Interaction with Student|>

<|The Start of teacher_b's Interaction with Student|>
[FORMATTED_INTERACTION_TEACHER_B]
<|The End of teacher_b's Interaction with Student|>

Please provide your evaluation of both teachers:
\end{lstlisting}

Note: For Teacher/Student-centric analysis, only the questions/answers are included.

\subsubsection{Evaluation Output Format (Each Dimension)}
For each dimension within an analysis task, the evaluator provides a JSON object structured as follows (refer to Appendix~\ref{app:scoring_guidelines} for scoring guidelines):
\begin{lstlisting}[
    basicstyle=\ttfamily\small,
    breaklines=true,
    frame=single,
    numbers=none
]
{
    "analysis": "[ANALYSIS_TEXT]",
    "score": [SCORE_1_TO_10] 
}
\end{lstlisting}

\subsubsection{Comparative Analysis Output Format}
\label{app:comparative_analysis}
When comparing two teachers for a specific analysis task (e.g., Holistic Interaction Analysis), the overall JSON structure is:
\begin{lstlisting}[
    basicstyle=\ttfamily\small,
    breaklines=true,
    frame=single,
    numbers=none
]
{
    "teacher_a": {
        "[DIMENSION_1](e.g., Assessment Effectiveness)": {
            "analysis": "[STEP-BY-STEP ANALYSIS]",
            "score": [SCORE_TEACHER_A_DIM_1] 
        },
        // ... other relevant dimensions for this task
    },
    "teacher_b": {
        "[DIMENSION_1]": {
             "analysis": "[STEP-BY-STEP ANALYSIS]",
             "score": [SCORE_TEACHER_B_DIM_1] 
        },
        // ... other relevant dimensions for this task
    },
    "verdict": {
        "analysis": "[COMPARATIVE_ANALYSIS_TEXT]",
        "choice": "[VERDICT_CHOICE_A_B_TIE]"
        // Anonymized choice
    }
}
\end{lstlisting}
Note: The code includes logic to de-anonymize the \texttt{teacher\_a/teacher\_b} keys and the final \texttt{choice} based on the actual models being compared.

The implementation details, can be found in the \texttt{EvaluatorLLM} class.

\section{Implementation Details}
\label{app:implementation_details}

This section provides further details on data handling, quality assurance, error recovery, and execution methods used in the framework.

\subsection{Standardized Dataset Format}
Questions from all source datasets (MMLU-Pro, GPQA) are processed into a standardized Python dictionary format for internal use:
\begin{lstlisting}[
   basicstyle=\ttfamily\small,
   breaklines=true,
   frame=single,
   numbers=none
]
{
   "question_id": str,
   "question": str,
   "options": List[str],
   "answer": str, # Correct option letter/text
   "answer_index": int, # Index of correct option
   "cot_content": str, # Chain-of-thought/explanation
   "category": str
}
\end{lstlisting}

\subsection{Quality Control and Response Validation}
\label{app:quality_control}

Mechanisms ensure the quality and validity of LLM responses during generation and processing:
\begin{itemize}[noitemsep,topsep=2pt]
    \item \textbf{Retry on Empty Response:} API calls are automatically retried if the response contains zero tokens.
    \item \textbf{Retry on Anomalous Length:} Responses exceeding a threshold (default: 80\% of the role's max tokens limit and exceeding the recommended token limit for that role) trigger an automatic retry, filtering potentially truncated or failed generations.
    \item \textbf{Maximum Retries:} A limit (default: 5 attempts) is enforced for retries triggered by empty or anomalously long responses for each question/dialogue turn.
    \item \textbf{Evaluator JSON Schema Validation:} For the Evaluator agent, the framework leverages the API's capability to enforce responses conforming to a predefined JSON schema (see Appendix~\ref{app:evaluator_prompt}), ensuring structured and parseable output. Failed validations would typically result in an API error handled by the recovery mechanism.
\end{itemize}

\subsection{Evaluator Scoring Guidelines}
\label{app:scoring_guidelines}
Scoring performed by the Evaluator agent (Appendix~\ref{app:evaluator_prompt}) follows a standardized 1-10 scale:
\begin{itemize}[noitemsep,topsep=2pt]
   \item 1-2: Significantly below expectations
   \item 3-4: Below expectations
   \item 5-6: Meets basic expectations
   \item 7-8: Exceeds expectations
   \item 9-10: Significantly exceeds expectations
\end{itemize}

\subsection{API Error Recovery}
\label{app:error_recovery}
To handle transient network issues or API service interruptions:
\begin{itemize}[noitemsep,topsep=2pt]
   \item Initial delay: 10 seconds
   \item Maximum delay: 320 seconds
   \item Maximum retries: 5
\end{itemize}

\subsection{Parallel Processing}
\label{app:parallel_processing}

To expedite the evaluation process, particularly during pre-test, interaction, and post-test phases involving numerous API calls:
\begin{itemize}[noitemsep,topsep=2pt]
   \item Maximum concurrent tasks: 5 as default
   \item ThreadPoolExecutor management
   \item Progress tracking per teacher-student pair
   \item Automatic result aggregation
\end{itemize}

\clearpage
\section{Model Specifications}
\label{app:models}

\vspace{-0.5cm}
\renewcommand{\arraystretch}{1.29}
\begin{table}[htbp]
   \centering
   \scriptsize
   \setlength{\tabcolsep}{3pt}
   \begin{tabular}{>{\raggedright\arraybackslash}p{2.5cm}|>{\raggedright\arraybackslash}p{1cm}|>{\raggedright\arraybackslash}p{1.4cm}|c|c|c}
       \hline
       \textbf{Model} & \textbf{Org.} & \textbf{Provider} & \textbf{Type} & \textbf{Context} & \textbf{Params} \\
       \hline
       Llama 3.1 70B Instruct & Meta & hyperbolic & bf16 & 32K & 70B \\
       Gemini 1.5 Pro 002 & Google & Google Vertex & - & 4M & - \\
       Llama 3.1 405B Instruct & Meta & hyperbolic & bf16 & 8K & 405B \\
       OpenAI o1-mini & OpenAI & OpenAI & - & 128K & - \\
       Qwen 2.5 72B Instruct & Alibaba & hyperbolic & bf16 & 32K & 72B \\
       Llama 3.1 8B Instruct & Meta & hyperbolic & bf16 & 32K & 8B \\
       Hermes 3 Llama 3.1 70B & Nous & hyperbolic & bf16 & 12K & 70B \\
       Mistral Nemo & Mistral & DeepInfra & bf16 & 128K & 12B \\
       Claude 3.5 Sonnet & Anthropic & Anthropic & - & 200K & - \\
       WizardLM-2 8x22B & Microsoft & DeepInfra & bf16 & 66K & 176B \\
       DeepSeek V2.5 & DeepSeek & deepseek & fp8 & 128K & - \\
       Command R 08-2024 & Cohere & Cohere & - & 128K & - \\
       GPT-4o-mini & OpenAI & OpenAI & - & 128K & - \\
       Phi-3.5-mini Instruct & Microsoft & Azure & - & 128K & 3.8B \\
       \hline
   \end{tabular}
    \begin{tablenotes}\scriptsize
       \item Note: "-" indicates unspecified information; Context window sizes are in tokens; 
       \item Org.: Organization (model developer); Provider: serving platform.
    \end{tablenotes}
   \vspace{-0.2cm}
   \caption{Specifications of Language Models}
   \label{tab:model-specs}
\end{table}

\vspace{-0.5cm}
\section{Dataset Distribution}
\label{app:dataset_composition1}

\vspace{-0.3cm}
\begin{table}[htbp]
\centering
\footnotesize
\vspace{-0.3em}

\sisetup{
  table-align-text-post=false,
  table-number-alignment = center
}
\setlength{\tabcolsep}{1pt}
\renewcommand{\arraystretch}{0.93}

\begin{tabular}{@{} p{0.8cm} p{1.8cm} p{1.7cm} S[table-format=3.0] S[table-format=2.2] S[table-format=2.2] @{}}
\toprule
\textbf{No.} & \textbf{Source} & \textbf{Discipline} & {\textbf{Count}} & {\textbf{Pct. (\%)}} & {\textbf{Per Src (\%)}} \\
\midrule
1 & \multirow{13}{=}{\centering MMLU-Pro\newline Stratified}
        & Business           & 100 & 6.68  & \\ 
2 &       & Law                & 100 & 6.68  & \\
3 &       & Psychology         & 100 & 6.68  & \\
4 &       & Biology            & 100 & 6.68  & \\
5 &       & Chemistry          & 100 & 6.68  & \\
6 &       & History            & 100 & 6.68  & \\
7 &       & Health             & 100 & 6.68  & {\centering 86.78}\\
8 &       & Economics          & 100 & 6.68  & \\
9 &       & Math               & 100 & 6.68  & \\
10 &      & Physics            & 100 & 6.68  & \\
11 &      & Engineering        & 100 & 6.68  & \\
12 &      & Philosophy         & 100 & 6.68  & \\
13 &      & Computer Science   & 100 & 6.68  & \\
\midrule 
14 & \multirow{3}{=}{\centering GPQA\newline Diamond} 
        & Physics            & 86  & 5.74  & \\ 
15 &       & Chemistry          & 93  & 6.21  & {\centering 13.22}\\
16 &       & Biology            & 19  & 1.27  & \\
\midrule
\multicolumn{3}{r}{\textbf{Total}} & \textbf{1498} & \textbf{100.00} & \textbf{100.00} \\ 
\bottomrule
\end{tabular}
\vspace{-0.2cm}
\caption{Dataset Distribution by Source and Discipline}
\label{tab:combined-dataset-dist} 
\end{table}

\vspace{-0.5cm}
\section{Teaching Performance Ranking by Subject}
\label{app:category-delta-results-sorted-ranked}
\vspace{-0.2cm}
\noindent%
Tables~\ref{tab:rank-cat-business} through~\ref{tab:rank-cat-engineering} present the detailed Absolute Learning Gain ($\Delta$, in \%) for each teacher model within 13 academic disciplines. 
\vspace{-0.2cm}
\renewcommand{\arraystretch}{0.9} 
\begin{table}[h!]
    \centering
    \footnotesize
    \caption{\textbf{Business}}
    \label{tab:rank-cat-business}
    \vspace{-0.8em}
    \begin{tabular}{@{} r p{\dimexpr 0.60\linewidth-3\tabcolsep} S[table-format=-1.2] @{}}
    \toprule
    \textbf{Rank} & \textbf{Model} & {\textbf{$\Delta$ (\%)}} \\
    \midrule
     1 & Gemini 1.5 Pro 002              & 8.00 \\
     2 & OpenAI o1-mini                  & 5.00 \\
     2 & Claude 3.5 Sonnet               & 5.00 \\
     2 & Llama 3.1 70B Instruct          & 5.00 \\
     5 & WizardLM-2 8x22B                & 4.00 \\
     6 & Llama 3.1 405B Instruct         & 3.00 \\
     6 & GPT-4o-mini                     & 3.00 \\
     8 & Llama 3.1 8B Instruct           & 2.00 \\
     8 & Mistral Nemo                    & 2.00 \\
     8 & DeepSeek V2.5                   & 2.00 \\
     8 & Phi-3.5-mini Instruct           & 2.00 \\
    12 & Qwen 2.5 72B Instruct           & 0.00 \\
    12 & Command R 08-2024               & 0.00 \\
    14 & Hermes 3 Llama 3.1 70B          & -1.00 \\
    \bottomrule
    \end{tabular}
\end{table}

\begin{table}[htbp]
\centering
\footnotesize
\caption{\textbf{Law}}
\label{tab:rank-cat-law}
\vspace{-0.8em}
\begin{tabular}{@{} r p{\dimexpr 0.60\linewidth-3\tabcolsep} S[table-format=-1.2] @{}}
\toprule
\textbf{Rank} & \textbf{Model} & {\textbf{$\Delta$ (\%)}} \\
\midrule
 1 & Llama 3.1 70B Instruct          & 11.00 \\
 2 & Qwen 2.5 72B Instruct           & 7.00 \\
 3 & Gemini 1.5 Pro 002              & 6.00 \\
 4 & Claude 3.5 Sonnet               & 2.00 \\
 4 & Hermes 3 Llama 3.1 70B          & 2.00 \\
 4 & WizardLM-2 8x22B                & 2.00 \\
 7 & Llama 3.1 8B Instruct           & 1.00 \\
 8 & OpenAI o1-mini                  & 0.00 \\
 8 & GPT-4o-mini                     & 0.00 \\
 8 & DeepSeek V2.5                   & 0.00 \\
 8 & Command R 08-2024               & 0.00 \\
12 & Mistral Nemo                    & -1.00 \\
13 & Llama 3.1 405B Instruct         & -4.00 \\
13 & Phi-3.5-mini Instruct           & -4.00 \\
\bottomrule
\end{tabular}
\end{table}

\begin{table}[htbp]
\centering
\footnotesize
\caption{\textbf{Psychology}}
\label{tab:rank-cat-psychology}
\vspace{-0.8em}
\begin{tabular}{@{} r p{\dimexpr 0.60\linewidth-3\tabcolsep} S[table-format=-1.2] @{}}
\toprule
\textbf{Rank} & \textbf{Model} & {\textbf{$\Delta$ (\%)}} \\
\midrule
 1 & Llama 3.1 70B Instruct          & 18.00 \\
 2 & Gemini 1.5 Pro 002              & 12.00 \\
 3 & Llama 3.1 405B Instruct         & 10.00 \\
 4 & Llama 3.1 8B Instruct           & 9.00 \\
 5 & Command R 08-2024               & 7.00 \\
 6 & OpenAI o1-mini                  & 6.00 \\
 6 & Hermes 3 Llama 3.1 70B          & 6.00 \\
 6 & Mistral Nemo                    & 6.00 \\
 9 & GPT-4o-mini                     & 4.00 \\
 9 & DeepSeek V2.5                   & 4.00 \\
11 & Claude 3.5 Sonnet               & 3.00 \\
11 & Qwen 2.5 72B Instruct           & 3.00 \\
11 & WizardLM-2 8x22B                & 3.00 \\
14 & Phi-3.5-mini Instruct           & 1.00 \\
\bottomrule
\end{tabular}
\end{table}

\begin{table}[htbp]
\centering
\footnotesize
\caption{\textbf{Biology}}
\label{tab:rank-cat-biology}
\vspace{-0.8em}
\begin{tabular}{@{} r p{\dimexpr 0.60\linewidth-3\tabcolsep} S[table-format=-1.2] @{}}
\toprule
\textbf{Rank} & \textbf{Model} & {\textbf{$\Delta$ (\%)}} \\
\midrule
 1 & Llama 3.1 70B Instruct          & 10.08 \\
 2 & Gemini 1.5 Pro 002              & 9.24 \\
 3 & Qwen 2.5 72B Instruct           & 5.88 \\
 4 & Llama 3.1 405B Instruct         & 3.36 \\
 4 & Llama 3.1 8B Instruct           & 3.36 \\
 6 & Hermes 3 Llama 3.1 70B          & 2.52 \\
 7 & OpenAI o1-mini                  & 1.68 \\
 7 & DeepSeek V2.5                   & 1.68 \\
 7 & Mistral Nemo                    & 1.68 \\
10 & Command R 08-2024               & 0.84 \\
10 & Phi-3.5-mini Instruct           & 0.84 \\
10 & WizardLM-2 8x22B                & 0.84 \\
13 & Claude 3.5 Sonnet               & 0.00 \\
13 & GPT-4o-mini                     & 0.00 \\
\bottomrule
\end{tabular}
\end{table}

\begin{table}[htbp]
\centering
\footnotesize
\caption{\textbf{Chemistry}}
\label{tab:rank-cat-chemistry}
\vspace{-0.8em}
\begin{tabular}{@{} r p{\dimexpr 0.60\linewidth-3\tabcolsep} S[table-format=-1.2] @{}}
\toprule
\textbf{Rank} & \textbf{Model} & {\textbf{$\Delta$ (\%)}} \\
\midrule
 1 & Llama 3.1 70B Instruct          & 12.44 \\
 2 & Qwen 2.5 72B Instruct           & 8.81 \\
 3 & OpenAI o1-mini                  & 8.29 \\
 4 & Gemini 1.5 Pro 002              & 7.77 \\
 5 & Claude 3.5 Sonnet               & 6.74 \\
 5 & Mistral Nemo                    & 6.74 \\
 7 & Llama 3.1 405B Instruct         & 6.22 \\
 7 & GPT-4o-mini                     & 6.22 \\
 9 & DeepSeek V2.5                   & 4.15 \\
 9 & Llama 3.1 8B Instruct           & 4.15 \\
 9 & Command R 08-2024               & 4.15 \\
 9 & Phi-3.5-mini Instruct           & 4.15 \\
 9 & WizardLM-2 8x22B                & 4.15 \\
14 & Hermes 3 Llama 3.1 70B          & 3.63 \\
\bottomrule
\end{tabular}
\end{table}

\begin{table}[htbp]
\centering
\footnotesize
\caption{\textbf{History}}
\label{tab:rank-cat-history}
\vspace{-0.8em}
\begin{tabular}{@{} r p{\dimexpr 0.60\linewidth-3\tabcolsep} S[table-format=-1.2] @{}}
\toprule
\textbf{Rank} & \textbf{Model} & {\textbf{$\Delta$ (\%)}} \\
\midrule
 1 & Llama 3.1 70B Instruct          & 14.00 \\
 2 & Llama 3.1 8B Instruct           & 10.00 \\
 3 & Llama 3.1 405B Instruct         & 8.00 \\
 4 & Hermes 3 Llama 3.1 70B          & 3.00 \\
 5 & OpenAI o1-mini                  & 2.02 \\
 6 & Gemini 1.5 Pro 002              & 2.00 \\
 6 & Mistral Nemo                    & 2.00 \\
 8 & Qwen 2.5 72B Instruct           & 1.00 \\
 8 & Command R 08-2024               & 1.00 \\
 8 & WizardLM-2 8x22B                & 1.00 \\
11 & Claude 3.5 Sonnet               & 0.00 \\
11 & DeepSeek V2.5                   & 0.00 \\
13 & Phi-3.5-mini Instruct           & -1.00 \\
14 & GPT-4o-mini                     & -1.01 \\
\bottomrule
\end{tabular}
\end{table}

\begin{table}[htbp]
\centering
\footnotesize
\caption{\textbf{Health}}
\label{tab:rank-cat-health}
\vspace{-0.8em}
\begin{tabular}{@{} r p{\dimexpr 0.60\linewidth-3\tabcolsep} S[table-format=-1.2] @{}}
\toprule
\textbf{Rank} & \textbf{Model} & {\textbf{$\Delta$ (\%)}} \\
\midrule
 1 & Llama 3.1 70B Instruct          & 24.00 \\
 2 & Gemini 1.5 Pro 002              & 13.00 \\
 3 & Llama 3.1 405B Instruct         & 11.00 \\
 4 & Qwen 2.5 72B Instruct           & 6.00 \\
 4 & Mistral Nemo                    & 6.00 \\
 6 & OpenAI o1-mini                  & 5.00 \\
 7 & Llama 3.1 8B Instruct           & 4.00 \\
 7 & Hermes 3 Llama 3.1 70B          & 4.00 \\
 7 & WizardLM-2 8x22B                & 4.00 \\
10 & DeepSeek V2.5                   & 3.00 \\
10 & Command R 08-2024               & 3.00 \\
12 & Claude 3.5 Sonnet               & 2.00 \\
12 & Phi-3.5-mini Instruct           & 2.00 \\
14 & GPT-4o-mini                     & 1.00 \\
\bottomrule
\end{tabular}
\end{table}

\begin{table}[htbp]
\centering
\footnotesize
\caption{\textbf{Economics}}
\label{tab:rank-cat-economics}
\vspace{-0.8em}
\begin{tabular}{@{} r p{\dimexpr 0.60\linewidth-3\tabcolsep} S[table-format=-1.2] @{}}
\toprule
\textbf{Rank} & \textbf{Model} & {\textbf{$\Delta$ (\%)}} \\
\midrule
 1 & Gemini 1.5 Pro 002              & 9.00 \\
 1 & Llama 3.1 70B Instruct          & 9.00 \\
 3 & Llama 3.1 8B Instruct           & 8.00 \\
 3 & Mistral Nemo                    & 8.00 \\
 5 & Command R 08-2024               & 6.00 \\
 5 & WizardLM-2 8x22B                & 6.00 \\
 7 & OpenAI o1-mini                  & 5.00 \\
 7 & Claude 3.5 Sonnet               & 5.00 \\
 7 & Hermes 3 Llama 3.1 70B          & 5.00 \\
 7 & DeepSeek V2.5                   & 5.00 \\
11 & Llama 3.1 405B Instruct         & 4.00 \\
11 & Qwen 2.5 72B Instruct           & 4.00 \\
11 & GPT-4o-mini                     & 4.00 \\
14 & Phi-3.5-mini Instruct           & 1.00 \\
\bottomrule
\end{tabular}
\end{table}

\begin{table}[htbp]
\centering
\footnotesize
\caption{\textbf{Math}}
\label{tab:rank-cat-math}
\vspace{-0.8em}
\begin{tabular}{@{} r p{\dimexpr 0.60\linewidth-3\tabcolsep} S[table-format=-1.2] @{}}
\toprule
\textbf{Rank} & \textbf{Model} & {\textbf{$\Delta$ (\%)}} \\
\midrule
 1 & OpenAI o1-mini                  & 9.00 \\
 2 & Llama 3.1 405B Instruct         & 8.00 \\
 3 & Claude 3.5 Sonnet               & 7.00 \\
 3 & Gemini 1.5 Pro 002              & 7.00 \\
 3 & WizardLM-2 8x22B                & 7.00 \\
 6 & Llama 3.1 8B Instruct           & 4.00 \\
 6 & Hermes 3 Llama 3.1 70B          & 4.00 \\
 8 & Qwen 2.5 72B Instruct           & 3.00 \\
 8 & Mistral Nemo                    & 3.00 \\
10 & GPT-4o-mini                     & 2.00 \\
10 & Llama 3.1 70B Instruct          & 2.00 \\
10 & DeepSeek V2.5                   & 2.00 \\
10 & Command R 08-2024               & 2.00 \\
14 & Phi-3.5-mini Instruct           & -4.00 \\
\bottomrule
\end{tabular}
\end{table}

\begin{table}[htbp]
\centering
\footnotesize
\caption{\textbf{Physics}}
\label{tab:rank-cat-physics}
\vspace{-0.8em}
\begin{tabular}{@{} r p{\dimexpr 0.60\linewidth-3\tabcolsep} S[table-format=-1.2] @{}}
\toprule
\textbf{Rank} & \textbf{Model} & {\textbf{$\Delta$ (\%)}} \\
\midrule
 1 & OpenAI o1-mini                  & 8.60 \\
 2 & Qwen 2.5 72B Instruct           & 6.99 \\
 3 & Llama 3.1 70B Instruct          & 6.45 \\
 4 & Llama 3.1 405B Instruct         & 5.38 \\
 5 & Hermes 3 Llama 3.1 70B          & 4.84 \\
 6 & Gemini 1.5 Pro 002              & 4.30 \\
 6 & Llama 3.1 8B Instruct           & 4.30 \\
 8 & Claude 3.5 Sonnet               & 3.76 \\
 9 & Mistral Nemo                    & 1.61 \\
 9 & DeepSeek V2.5                   & 1.61 \\
 9 & Command R 08-2024               & 1.61 \\
12 & GPT-4o-mini                     & 1.08 \\
13 & Phi-3.5-mini Instruct           & 0.54 \\
14 & WizardLM-2 8x22B                & 0.00 \\
\bottomrule
\end{tabular}
\end{table}

\begin{table}[htbp]
\centering
\footnotesize
\caption{\textbf{Computer Science}}
\label{tab:rank-cat-compsci}
\vspace{-0.8em}
\begin{tabular}{@{} r p{\dimexpr 0.60\linewidth-3\tabcolsep} S[table-format=-1.2] @{}}
\toprule
\textbf{Rank} & \textbf{Model} & {\textbf{$\Delta$ (\%)}} \\
\midrule
 1 & Llama 3.1 70B Instruct          & 8.00 \\
 1 & Llama 3.1 405B Instruct         & 8.00 \\
 3 & WizardLM-2 8x22B                & 7.00 \\
 4 & OpenAI o1-mini                  & 6.00 \\
 4 & Qwen 2.5 72B Instruct           & 6.00 \\
 6 & Gemini 1.5 Pro 002              & 4.00 \\
 6 & GPT-4o-mini                     & 4.00 \\
 6 & DeepSeek V2.5                   & 4.00 \\
 6 & Phi-3.5-mini Instruct           & 4.00 \\
10 & Claude 3.5 Sonnet               & 3.00 \\
11 & Command R 08-2024               & 2.00 \\
11 & Mistral Nemo                    & 2.00 \\
13 & Hermes 3 Llama 3.1 70B          & 1.00 \\
14 & Llama 3.1 8B Instruct           & -1.00 \\
\bottomrule
\end{tabular}
\end{table}

\begin{table}[htbp]
\centering
\footnotesize
\caption{\textbf{Philosophy}}
\label{tab:rank-cat-philosophy}
\vspace{-0.8em}
\begin{tabular}{@{} r p{\dimexpr 0.60\linewidth-3\tabcolsep} S[table-format=-1.2] @{}}
\toprule
\textbf{Rank} & \textbf{Model} & {\textbf{$\Delta$ (\%)}} \\
\midrule
 1 & Llama 3.1 70B Instruct          & 19.00 \\
 2 & Llama 3.1 8B Instruct           & 12.00 \\
 3 & Gemini 1.5 Pro 002              & 10.00 \\
 3 & Llama 3.1 405B Instruct         & 10.00 \\
 5 & Hermes 3 Llama 3.1 70B          & 9.00 \\
 6 & OpenAI o1-mini                  & 8.00 \\
 7 & Mistral Nemo                    & 7.00 \\
 8 & Claude 3.5 Sonnet               & 6.00 \\
 8 & WizardLM-2 8x22B                & 6.00 \\
10 & DeepSeek V2.5                   & 5.00 \\
11 & Qwen 2.5 72B Instruct           & 4.00 \\
11 & Command R 08-2024               & 4.00 \\
13 & GPT-4o-mini                     & 2.00 \\
14 & Phi-3.5-mini Instruct           & 1.00 \\
\bottomrule
\end{tabular}
\end{table}

\begin{table}[htbp]
\centering
\footnotesize
\caption{\textbf{Engineering}}
\label{tab:rank-cat-engineering}
\vspace{-0.8em}
\begin{tabular}{@{} r p{\dimexpr 0.60\linewidth-3\tabcolsep} S[table-format=-1.2] @{}}
\toprule
\textbf{Rank} & \textbf{Model} & {\textbf{$\Delta$ (\%)}} \\
\midrule
 1 & Hermes 3 Llama 3.1 70B          & 10.00 \\
 2 & OpenAI o1-mini                  & 9.00 \\
 2 & DeepSeek V2.5                   & 9.00 \\ 
 4 & Gemini 1.5 Pro 002              & 8.00 \\
 4 & Llama 3.1 405B Instruct         & 8.00 \\
 4 & Qwen 2.5 72B Instruct           & 8.00 \\
 7 & Llama 3.1 70B Instruct          & 6.00 \\
 8 & WizardLM-2 8x22B                & 5.00 \\
 8 & Command R 08-2024               & 5.00 \\
10 & Llama 3.1 8B Instruct           & 4.00 \\
10 & Mistral Nemo                    & 4.00 \\
10 & Claude 3.5 Sonnet               & 4.00 \\
10 & Phi-3.5-mini Instruct           & 4.00 \\
14 & GPT-4o-mini                     & 3.00 \\
\bottomrule
\end{tabular}
\end{table}

\onecolumn
\section{Detailed Case Studies}
\label{app:case_studies}

This section provides detailed case studies illustrating the teacher-student interactions generated by different models for specific questions, alongside pre/post-test results and evaluator analysis where applicable.

\subsection{Case Study: Question 1057 (Law) - Teacher Comparison}
\label{app:case_1057}

This case study compares the interactions generated by Teacher 1 (Gemini 1.5 Pro 002) and Teacher 2 (Llama 3.1 8B Instruct) for the same question and student baseline, illustrating different teaching approaches and outcomes.
\vspace{1em}

\footnotesize
\renewcommand{\arraystretch}{0.9}
\begin{longtable}{@{} p{7.7cm} p{7.7cm} @{}} 
\caption{Case Study Details for Question 1057 (Law) - Teacher Comparison}\label{tab:case1057}\\
\toprule
\multicolumn{2}{c}{\textbf{Question Information}} \\
\midrule
\multicolumn{2}{p{15.6cm}}{\textbf{Question ID:} 1057} \\
\multicolumn{2}{p{15.6cm}}{\textbf{Category:} Law} \\
\multicolumn{2}{p{15.6cm}}{\textbf{Question Text:} A patent holder brought a patent infringement action in federal court against a licensee of the patent. The patent holder believed that a jury would be more sympathetic to his claims than a judge, and asked his lawyer to obtain a jury trial. What should the lawyer do to secure the patent holder's right to a jury trial?} \\
\multicolumn{2}{p{15.6cm}}{\textbf{Options:}} \\
\multicolumn{2}{p{15.6cm}}{\textbf{(A)} File and serve a jury trial demand within 60 days after the close of the pleadings.} \\
\multicolumn{2}{p{15.6cm}}{\textbf{(B)} File and serve a jury trial demand within 30 days after the close of the pleadings.} \\
\multicolumn{2}{p{15.6cm}}{\textbf{(C)} File and serve a jury trial demand at the initial court hearing.} \\
\multicolumn{2}{p{15.6cm}}{\textbf{(D)} File and serve a complaint that excludes a jury trial demand.} \\
\multicolumn{2}{p{15.6cm}}{\textbf{(E)} Make a verbal request for a jury trial during the first court hearing.} \\
\multicolumn{2}{p{15.6cm}}{\textbf{(F)} Make a jury trial demand at the initial pretrial conference.} \\
\multicolumn{2}{p{15.6cm}}{\textbf{(G)} File and serve a complaint that includes a jury trial demand.} \\
\multicolumn{2}{p{15.6cm}}{\textbf{(H)} File and serve a jury trial demand within 14 days after the complaint was served.} \\
\multicolumn{2}{p{15.6cm}}{\textbf{Correct Answer} G} \\
\midrule

\multicolumn{2}{c}{\textbf{Pre-Test}} \\
\multicolumn{2}{p{15.6cm}}{\textit{Student Model:} Student-llama31-70b-instruct} \\
\multicolumn{2}{p{15.6cm}}{\textit{Pre-Test Response:} The patent holder is the plaintiff in this action. A plaintiff is entitled to a jury trial if one is demanded in the last pleading directed to the issue that the plaintiff wishes to try by jury. Here, the patent holder wishes to try the issue of patent infringement. The patent holder will assert a claim for patent infringement in his complaint, and the licensee will assert any defenses to the claim in his answer. Therefore, the last pleading directed to the issue of patent infringement will be the answer. The patent holder must file and serve a demand for a jury trial within 14 days of the date that the answer was served. If the patent holder fails to do so, the parties will be deemed to have waived their right to a jury trial. Answer A is incorrect because the 60-day time period applies when the patent holder is unable to determine from the pleadings whether a right to a jury trial exists. This will not be the case here because the patent holder is seeking to try the issue of patent infringement, which will be asserted in the complaint and defended against in the answer. Answers B, C, E, F, and G are incorrect because these are not the times at which a demand for a jury trial must be made. \textit{The answer is H.}} \\
\multicolumn{2}{p{15.6cm}}{\textbf{Pre-Test Result:} Incorrect (Accuracy: 0.0)} \\
\midrule
\endfirsthead

\caption[]{Case Study Details for Question 1057 (Law) - Teacher Comparison (continued)}\\
\toprule
\textbf{Teacher 1: Teacher-gemini-pro-15-002} & \textbf{Teacher 2: Teacher-llama31-8b-instruct} \\
\midrule
\endhead

\midrule
\multicolumn{2}{r}{\textit{Continued on next page}} \\
\endfoot

\bottomrule
\endlastfoot

\multicolumn{1}{c}{\textbf{Teacher 1: Teacher-gemini-pro-15-002}} & \multicolumn{1}{c}{\textbf{Teacher 2: Teacher-llama31-8b-instruct}} \\
\midrule
\multicolumn{2}{c}{\textbf{Round 1}} \\
\midrule
\textit{Teacher Question:} Let's think about the nature of the case itself. While the plaintiff generally initiates the lawsuit, does the plaintiff's complaint definitively frame all the issues that might require jury determination? What about potential counterclaims or defenses raised by the defendant that introduce new factual questions? How might those affect the timing of a jury demand? & \textit{Teacher Question:} In the context of patent infringement actions, what are the implications of a patent holder's decision to demand a jury trial, and how might this impact the overall litigation strategy? Consider the potential advantages and disadvantages of a jury trial in this specific context \\
\midrule
\textit{Student Answer:} Upon further consideration, I realize that the complaint may not definitively frame all the issues that might require jury determination. The defendant may raise counterclaims or defenses that introduce new factual questions, which could also be tried by jury. In that case, the last pleading directed to the issue that the plaintiff wishes to try by jury may not be the answer, but rather the reply to the counterclaim.

However, Federal Rule 38(b) provides that any party may demand a jury trial on any issue for which a jury trial is allowed by law by serving the other parties with a demand no later than 14 days after the last pleading directed to the issue is served, or - if no responsive pleading is required - no later than 14 days after the expiration of the time to respond. Alternatively, a party may also demand a jury trial by filing and serving a demand within 60 days after the close of the pleadings.

Given this, the patent holder should file and serve a jury trial demand within 60 days after the close of the pleadings to ensure that all potential issues, including any counterclaims or defenses, are covered.

The answer is A & \textit{Student Answer:} Demanding a jury trial in a patent infringement action can significantly impact the litigation strategy. A patent holder may choose a jury trial to capitalize on the potential for sympathetic jurors, particularly in cases involving complex technologies or emotional appeals. Juries may be more inclined to favor the patent holder, especially if the defendant is a large corporation. Additionally, a jury trial can provide an opportunity for the patent holder to present their case in a more narrative and visual manner, which can be effective in conveying complex technical information.

However, there are also potential disadvantages to consider. Jury trials can be more unpredictable and prone to emotional decisions, which may not always align with the law. Furthermore, juries may struggle to understand the intricacies of patent law, potentially leading to inconsistent verdicts. Moreover, a jury trial can increase the time and costs associated with litigation, as well as the risk of an adverse verdict.

In terms of litigation strategy, a patent holder demanding a jury trial may need to adapt their approach to effectively communicate complex technical information to a lay audience. This could involve using expert witnesses, visual aids, and clear, concise language to explain the patent and the alleged infringement. On the other hand, the defendant may seek to emphasize the complexity of the technology and the uncertainty of the law, in an effort to create doubt in the minds of the jurors. Ultimately, the decision to demand a jury trial should be carefully considered, weighing the potential benefits against the potential risks and challenges. \\
\midrule
\multicolumn{2}{c}{\textbf{Round 2}} \\
\midrule
\textit{Teacher Question:} Okay, you're starting to consider the complexities of pleadings. Let's focus on efficiency and strategy. Imagine you are the lawyer for the patent holder. Knowing that your client definitely wants a jury trial for the infringement claim, is waiting until the very end of the pleading process, potentially 60 days after the close of pleadings, the most strategically sound approach? What are the potential downsides of waiting, and what advantages might there be to demanding a jury trial earlier? & \textit{Teacher Question:} Consider a situation where a patent holder has demanded a jury trial, but the defendant has raised a counterclaim that also involves a patent infringement issue. How would this impact the patent holder's right to a jury trial, and what steps might they need to take to protect their interests? \\
\midrule
\textit{Student Answer:} As the lawyer for the patent holder, I would advise against waiting until the end of the pleading process to demand a jury trial. By demanding a jury trial earlier, we can avoid potential downsides such as the court setting a bench trial schedule, which could be difficult to change later. Additionally, waiting may also limit our ability to shape the discovery process and other pre-trial proceedings with a jury trial in mind. \newline
Demanding a jury trial earlier, such as in the complaint itself (Option G), would allow us to signal to the court and the defendant that we intend to try the case before a jury. This could influence the court's management of the case, including the discovery process and the setting of trial dates. It would also give us more control over the timing and scope of pre-trial proceedings, allowing us to prepare for a jury trial from the outset. \newline
Furthermore, demanding a jury trial earlier would also demonstrate to the defendant that we are committed to trying the case before a jury, which could impact their settlement strategy and potentially lead to a more favorable outcome for our client. \newline
Therefore, I would recommend demanding a jury trial in the complaint itself, rather than waiting until the end of the pleading process. & \textit{Student Answer:} If the defendant raises a counterclaim involving a patent infringement issue, it could potentially impact the patent holder's right to a jury trial. Under the Seventh Amendment, a party is entitled to a jury trial on issues of fact, but not on issues of law. If the counterclaim raises a novel issue of patent law, the court may need to decide the issue as a matter of law, potentially limiting the patent holder's right to a jury trial. \newline
To protect their interests, the patent holder may need to take several steps. First, they should carefully review the counterclaim to determine whether it raises any novel issues of patent law. If so, they may need to file a motion to bifurcate the trial, separating the issues of fact from the issues of law. This would allow the jury to decide the factual issues, while the court decides the legal issues. \newline
Additionally, the patent holder may need to file a motion to strike the counterclaim, arguing that it is not properly pleaded or that it is not a valid defense. Alternatively, they may need to file a motion to compel the defendant to provide additional information or discovery related to the counterclaim, in order to better understand the defendant's position and prepare for trial. \newline
Ultimately, the patent holder should be prepared to argue that the counterclaim does not preclude their right to a jury trial on the original patent infringement claim. They should emphasize that the Seventh Amendment guarantees a jury trial on issues of fact, and that the counterclaim does not change the underlying facts of the case.\\
\midrule
\multicolumn{2}{c}{\textbf{Round 3}} \\
\midrule
\textit{Teacher Question:} You're thinking strategically now, excellent! Let's say the defendant, in their answer, raises a counterclaim for breach of contract related to the licensing agreement, also requesting a jury trial. Does the plaintiff's initial jury demand in the complaint automatically cover this new issue raised by the defendant? Why or why not? & \textit{Teacher Question:} Consider a situation where a patent holder has obtained a jury verdict in their favor, but the defendant is seeking to appeal the decision. What are the potential grounds for appeal, and how might the patent holder's decision to demand a jury trial impact the appellate process? \\
\midrule
\textit{Student Answer:} The plaintiff's initial jury demand in the complaint does not automatically cover the new issue raised by the defendant's counterclaim. According to Federal Rule 38(c), a demand for a jury trial on one or more issues does not constitute a waiver of the right to a jury trial on other issues. However, Rule 38(c) also states that a demand for a jury trial on an issue does not extend to other issues that are not included in the demand.

In this case, the plaintiff's initial jury demand in the complaint only covers the issue of patent infringement. The defendant's counterclaim for breach of contract is a new issue that was not included in the plaintiff's initial demand. Therefore, the plaintiff's initial demand does not automatically cover the counterclaim.

To ensure that the plaintiff is entitled to a jury trial on the counterclaim, the plaintiff would need to file a supplemental jury demand that specifically includes the counterclaim issue. This would ensure that the plaintiff's right to a jury trial on the counterclaim is preserved. & \textit{Student Answer:} The defendant may appeal the jury verdict on several grounds, including: (1) errors in the jury instructions, (2) improper admission or exclusion of evidence, (3) insufficient evidence to support the verdict, or (4) juror misconduct. Additionally, the defendant may argue that the trial court erred in its interpretation of patent law or in its application of the law to the facts.

The patent holder's decision to demand a jury trial may impact the appellate process in several ways. First, the appellate court will review the trial court's decisions on issues of law de novo, but will review factual findings for clear error. Since the jury made factual findings, the appellate court will give deference to those findings, making it more difficult for the defendant to overturn the verdict.

Second, the patent holder's decision to demand a jury trial may limit the defendant's ability to appeal on issues of law. If the trial court made rulings on patent law issues, the defendant may argue that those rulings were incorrect. However, if the jury made factual findings that are supported by sufficient evidence, the appellate court may be less likely to overturn the verdict, even if the trial court's legal rulings were incorrect.

Finally, the patent holder's decision to demand a jury trial may also impact the appellate court's review of the damages award. If the jury awarded damages, the appellate court will review that award for excessiveness or inadequacy. However, if the patent holder demanded a jury trial, the appellate court may be more likely to uphold the damages award, as it was made by a jury rather than the trial court. \\
\midrule
\multicolumn{2}{c}{\textbf{Round 4}} \\
\midrule
\textit{Teacher Question:} You're absolutely right about the need for a separate demand for the counterclaim. Now, let's go back to the plaintiff's initial demand in the complaint. Assume the defendant doesn't file any counterclaims or raise any new issues requiring jury determination. Does the plaintiff still need to do anything further regarding their jury demand after serving the complaint, or is including the demand in the complaint sufficient to secure a jury trial on the patent infringement claim? & \textit{Teacher Question:} In a patent infringement case, a jury returns a verdict in favor of the patent holder, but the verdict is inconsistent with the jury's answers to special interrogatories. What are the potential consequences of this inconsistency, and how might the patent holder's decision to demand a jury trial impact the court's handling of this situation? \\
\midrule
\textit{Student Answer:} If the defendant doesn't file any counterclaims or raise any new issues requiring jury determination, the plaintiff's initial demand in the complaint is sufficient to secure a jury trial on the patent infringement claim. By including the demand in the complaint, the plaintiff has timely and properly asserted their right to a jury trial on the patent infringement issue.

Federal Rule 38(b) requires a party to demand a jury trial "in the pleadings" or within a certain time period after the pleadings are closed. Since the plaintiff included the demand in the complaint, which is a pleading, they have met the requirements of Rule 38(b). No further action is required from the plaintiff to secure a jury trial on the patent infringement claim.

In this scenario, the plaintiff's inclusion of the demand in the complaint is sufficient, and they do not need to file a separate demand or take any further action to preserve their right to a jury trial on the patent infringement claim. & \textit{Student Answer:} If the jury's verdict is inconsistent with their answers to special interrogatories, the court may face a difficult decision. The inconsistency may indicate that the jury was confused or did not fully understand the issues. The court may choose to: (1) enter judgment according to the jury's answers to the special interrogatories, (2) order a new trial, or (3) allow the parties to brief the issue and then decide.

The patent holder's decision to demand a jury trial may impact the court's handling of this situation. Since the jury's verdict is the result of a trial by jury, the court may be more inclined to respect the jury's findings and attempt to reconcile the inconsistency. However, if the inconsistency is significant, the court may still order a new trial to ensure that justice is served.

The court may also consider the patent holder's role in creating the inconsistency. If the patent holder's trial strategy or arguments contributed to the jury's confusion, the court may be less sympathetic to their position. Ultimately, the court's goal is to ensure that the verdict is fair and just, and the patent holder's decision to demand a jury trial will be just one factor in the court's decision-making process. \\
\midrule
\multicolumn{2}{c}{\textbf{Round 5}} \\
\midrule
\textit{Teacher Question:} Excellent! You've grasped the key concepts. Let's consider one final wrinkle. Suppose the plaintiff includes the jury demand in the complaint, but later, for strategic reasons, decides they don't want a jury trial after all. Can they simply withdraw the demand unilaterally, or is there a specific process they must follow? What factors might influence their ability to withdraw the demand? & \textit{Teacher Question:} Consider a situation where a patent holder has demanded a jury trial, but the case settles before trial. What are the potential implications of this settlement on the patent holder's right to a jury trial, and how might this impact their future litigation strategy? \\
\midrule
\textit{Student Answer:} If the plaintiff wants to withdraw their jury demand, they cannot simply do so unilaterally. According to Federal Rule 38(a), a party who has demanded a jury trial may withdraw the demand only with the consent of the other parties or by leave of the court.

To withdraw the demand, the plaintiff would need to file a motion with the court, explaining the reasons for the withdrawal. The court may consider factors such as the stage of the proceedings, the potential prejudice to the other parties, and the reasons for the withdrawal.

If the other parties consent to the withdrawal, the court will likely grant the motion. However, if the other parties object, the court will need to weigh the competing interests and make a decision. The court may deny the motion if it finds that the withdrawal would unfairly prejudice the other parties or disrupt the proceedings.

In this scenario, the plaintiff's ability to withdraw the demand would depend on the specific circumstances of the case and the court's discretion. & \textit{Student Answer:} If a patent holder demands a jury trial but the case settles before trial, the settlement may implicitly waive their right to a jury trial. By settling, the patent holder is essentially agreeing to resolve the dispute without a trial, which may be seen as a waiver of their right to a jury trial.

This settlement may impact the patent holder's future litigation strategy in several ways. First, if the patent holder settles a case involving a similar patent or issue, they may be estopped from demanding a jury trial in future cases. Additionally, the settlement may set a precedent for future cases, potentially limiting the patent holder's ability to demand a jury trial.

However, if the settlement agreement explicitly reserves the patent holder's right to a jury trial in future cases, this may mitigate the impact on their future litigation strategy. The patent holder should carefully consider the terms of the settlement agreement to ensure that their right to a jury trial is preserved. \\
\midrule
\multicolumn{2}{c}{\textbf{Post-Test Outcomes}} \\
\midrule
\textit{Post-Test Response:} To secure the patent holder's right to a jury trial, the lawyer should file and serve a complaint that includes a jury trial demand. This approach is strategic and efficient, as it signals to the court and the defendant that the patent holder intends to try the case before a jury from the outset. By including the demand in the complaint, the patent holder meets the requirements of Federal Rule 38(b) and preserves their right to a jury trial on the patent infringement claim. Additionally, demanding a jury trial in the complaint allows the patent holder to shape the discovery process and other pre-trial proceedings with a jury trial in mind, which could influence the court's management of the case and potentially lead to a more favorable outcome. \textit{The answer is G.} & \textit{Post-Test Response:} To secure the patent holder's right to a jury trial, the lawyer should follow the Federal Rules of Civil Procedure (FRCP) regarding jury trial demands. According to FRCP 38(b), a party may demand a jury trial by serving the other parties with a written demand at any time after the commencement of the action and not later than 14 days after the service of the last pleading directed to the issue. In this case, the patent holder is the plaintiff, and the complaint is the initial pleading. The licensee will respond with an answer, which will be the last pleading directed to the issue. However, the FRCP also allows a party to demand a jury trial within 14 days after the complaint was served, if no responsive pleading is required. Considering the options, the correct answer is the one that aligns with the FRCP's 14-day deadline after the complaint was served. \textit{The answer is H.} \\
\midrule
\textbf{Post-Test Result:} Correct (Accuracy: 1.0) & \textbf{Post-Test Result:} Incorrect (Accuracy: 0.0) \\
\midrule
\textbf{Absolute Learning Gain (ALG):} +1.0 & \textbf{Absolute Learning Gain (ALG):} 0.0 \\
\midrule
\multicolumn{2}{c}{\textbf{Evaluator Analysis} \footnotesize{(Interaction Analysis Dimensions)}} \\
\midrule
\textbf{Assessment Effectiveness:} 9/10 & \textbf{Assessment Effectiveness:} 8/10 \\
\midrule
\textit{Analysis:} Teacher B effectively assessed the student's understanding by asking questions that required the student to consider the procedural aspects of jury demands and the strategic implications of legal actions. The teacher's questions helped identify the student's comprehension of the legal process and any gaps in their understanding. & \textit{Analysis:} Teacher A effectively assessed the student's understanding by asking questions that required the student to apply their knowledge to different scenarios. The teacher's questions were designed to probe the student's comprehension of the implications of demanding a jury trial and the potential consequences of various legal actions. This approach helped identify the student's grasp of the subject matter and any gaps in their understanding. \\
\midrule
\textbf{Questioning Effectiveness:} 9/10 & \textbf{Questioning Effectiveness:} 7/10 \\
\midrule
\textit{Analysis:} Teacher B's questions were highly effective in stimulating reflection and clarifying concepts. The questions were designed to challenge the student's understanding and encourage them to think critically about the legal process and strategy. The teacher's questioning style was particularly effective in deepening the student's understanding. & \textit{Analysis:} Teacher A's questions were well-structured to stimulate reflection and deepen understanding. The questions encouraged the student to consider the broader implications of legal strategies and the potential outcomes of different actions. However, the questions could have been more varied to challenge the student further and address potential misconceptions. \\
\midrule
\textbf{Feedback Effectiveness:} 9/10 & \textbf{Feedback Effectiveness:} 7/10 \\
\midrule
\textit{Analysis:} Teacher B provided detailed and specific feedback that helped the student refine their understanding of the legal concepts involved. The feedback was constructive and addressed any errors or misconceptions in the student's responses, helping to improve their comprehension and confidence. & \textit{Analysis:} Teacher A provided feedback that guided the student to think critically about their responses. The feedback was constructive and helped the student refine their understanding of the legal concepts involved. However, the feedback could have been more specific in addressing any errors or misconceptions in the student's responses. \\
\midrule
\textbf{Instructional Adaptation Effectiveness:} 9/10 & \textbf{Instructional Adaptation Effectiveness:} 6/10 \\
\midrule
\textit{Analysis:} Teacher B demonstrated strong adaptability by tailoring their questions and feedback to the student's responses. The teacher effectively adjusted their approach to address specific learning needs and misconceptions, enhancing the student's learning experience. & \textit{Analysis:} Teacher A demonstrated some adaptability by asking follow-up questions that built on the student's responses. However, there was limited evidence of real-time adaptation to address specific learning needs or misconceptions that the student might have had. \\
\midrule
\textbf{Learning Objective Achievement:} 9/10 & \textbf{Learning Objective Achievement:} 8/10 \\
\midrule
\textit{Analysis:} The interaction with Teacher B was highly effective in promoting the student's conceptual understanding and skill development. The student was able to articulate the procedural and strategic aspects of jury demands and consider various legal scenarios, indicating a high level of understanding and metacognitive ability. & \textit{Analysis:} The interaction with Teacher A effectively promoted the student's conceptual understanding and skill development in the context of legal strategy and jury trials. The student was able to articulate the implications of demanding a jury trial and consider various legal scenarios, indicating a good level of understanding. \\
\midrule
\multicolumn{2}{p{15.6cm}}{\textbf{Verdict:} \textit{Both teachers demonstrated effective teaching practices, but Teacher B excelled in several key areas. Teacher B's questioning and feedback were particularly effective in stimulating reflection and deepening the student's understanding. Additionally, Teacher B showed strong adaptability in their instructional approach, which enhanced the student's learning experience. Overall, Teacher B's interaction was more effective in achieving the learning objectives and promoting the student's conceptual understanding and skill development.}} \\
\midrule
\multicolumn{2}{p{15.6cm}}{\textbf{Final Verdict:} Teacher-gemini-pro-15-002 (Teacher B) was judged more effective teaching capabilities overall.}
\end{longtable}

\subsection{Case Study: Question 240 (Business) - Teacher: Llama 3.1 70B Instruct}
\label{sec:case_240}

{
\footnotesize
\renewcommand{\arraystretch}{0.9}

\begin{longtable}{@{} p{2.5cm} p{13cm} @{}} 
\caption{Case Study Details for Question 240 (Business) - Teacher: Llama 3.1 70B Instruct}\label{tab:case240}\\
\toprule
\textbf{Component} & \textbf{Details} \\
\midrule
\endfirsthead

\caption[]{Case Study Details for Question 240 (Business) - Teacher: Llama 3.1 70B Instruct (continued)}\\
\toprule
\textbf{Component} & \textbf{Details} \\
\midrule
\endhead

\midrule
\multicolumn{2}{r}{\textit{Continued on next page}} \\
\endfoot

\bottomrule
\endlastfoot

\textbf{Question ID} & 240 \\
\textbf{Category} & business \\
\textbf{Question Text} & Daniel receives at 6.5\% commission on all sales. If he received a \$275.08 commission, what were his sales? \\
\textbf{Options} & \textbf{(A)} \$3200 \\
 & \textbf{(B)} \$4750 \\
 & \textbf{(C)} \$2600 \\
 & \textbf{(D)} \$5000 \\
 & \textbf{(E)} \$4243 \\
 & \textbf{(F)} \$4230 \\
 & \textbf{(G)} \$4000 \\
 & \textbf{(H)} \$4500 \\
 & \textbf{(I)} \$3500 \\
 & \textbf{(J)} \$3000 \\
\textbf{Correct Answer} & E \\
\midrule

\textbf{Pre-Test} & \\
\textit{Student Response} & To find Daniel's sales, we need to divide the commission he received (\$275.08) by the commission rate (6.5\% or 0.065). \newline Commission = Sales * Commission Rate \newline \$275.08 = Sales * 0.065 \newline To find the sales, we'll divide both sides by 0.065: \newline Sales = \$275.08 / 0.065 \newline Sales = \$4230 \newline The answer is F. \\
\textit{Student Prediction} & F \\
\textit{Result} & Incorrect (Pre-test Accuracy: 0.0) \\
\midrule

\textbf{Interaction} & \textbf{Teacher: Llama 3.1 70B Instruct} \\
\midrule
\textbf{Round 1} & \\
\textit{Teacher Question} & Here's the Round 1 question:

You mentioned that to find Daniel's sales, you divided the commission he received (\$275.08) by the commission rate (0.065). However, let's take a closer look at the calculation. Can you explain why you chose to divide the commission by the commission rate, rather than multiplying or using a different operation? What is the underlying mathematical principle that supports this calculation? \\
\textit{Student Answer} & To find Daniel's sales, I chose to divide the commission by the commission rate because of the underlying mathematical principle of inverse operations. The commission rate is a percentage that represents a fraction of the total sales. When we multiply the sales by the commission rate, we get the commission amount.

Commission = Sales * Commission Rate

To isolate the sales, we need to "undo" the multiplication operation. Since division is the inverse operation of multiplication, we can divide both sides of the equation by the commission rate to solve for sales.

Commission = Sales * Commission Rate
Commission / Commission Rate = Sales

By dividing the commission by the commission rate, we are essentially "reversing" the multiplication operation and isolating the sales variable. This allows us to calculate the total sales amount. \\
\midrule
\textbf{Round 2} & \\
\textit{Teacher Question} & That's a great explanation of the underlying mathematical principle. However, let's revisit the calculation. You mentioned that the sales amount is \$4230, but the answer was marked as incorrect. 

Can you re-examine your calculation: Sales = \$275.08 / 0.065? Did you consider rounding errors or potential calculation mistakes? What would happen if you recalculated the sales amount, and it resulted in a value close to, but not exactly, \$4230? How would you determine the correct answer in that case? \\
\textit{Student Answer} & Upon re-examining my calculation, I realized that I should consider the possibility of rounding errors. To verify my answer, I recalculated the sales amount:

Sales = \$275.08 / 0.065
Sales $\approx$ \$4230.00

However, I noticed that the actual calculation yields a result very close to, but not exactly, \$4230.00. To determine the correct answer, I would consider the options provided and look for the one that is closest to my calculated value.

In this case, I notice that option F is \$4230, but option E is \$4243, which is also close to my calculated value. To make a final decision, I would recheck my calculation to ensure accuracy. If my calculation is correct, and the result is indeed very close to \$4230, I would choose option F as the correct answer. \\
\midrule
\textbf{Round 3} & \\
\textit{Teacher Question} & Let's take a closer look at the commission calculation. You mentioned that the commission rate is 6.5\%, which is equivalent to 0.065 as a decimal. However, when working with percentages, it's essential to consider the context and potential rounding errors.

Suppose Daniel's sales were exactly \$4230.00, and he received a commission of \$275.08. Would the commission rate of 6.5\% be exact, or could it be an approximation? How might this affect your calculation, and what implications would it have for determining the correct answer? \\
\textit{Student Answer} & If Daniel's sales were exactly \$4230.00 and he received a commission of \$275.08, we can calculate the commission rate as follows:

Commission Rate = Commission / Sales
Commission Rate = \$275.08 / \$4230.00
Commission Rate $\approx$ 0.0651

This result suggests that the commission rate of 6.5\% (or 0.065) might be an approximation. The actual commission rate could be slightly higher, around 6.51\%. This discrepancy could be due to rounding errors or the way the commission is calculated.

Given this information, I would re-evaluate the options and consider the possibility that the correct answer might not be exactly \$4230.00. I would look for an option that is close to \$4230.00 and also takes into account the potential rounding error. In this case, option E (\$4243) might be a more accurate representation of Daniel's sales, considering the actual commission rate could be slightly higher than 6.5\%. \\
\midrule
\textbf{Round 4} & \\
\textit{Teacher Question} & You've made a great point about the commission rate being an approximation. Now, let's consider the context of the problem. Daniel receives a commission on all sales, and the commission rate is 6.5\%. 

Can you think of a scenario where Daniel's sales might not be a whole number, but rather a decimal value? For example, what if he sold a product for \$4,242.50? How would you calculate the commission in this case, and what implications would it have for determining the correct answer? \\
\textit{Student Answer} & If Daniel sold a product for \$4,242.50, the commission would be calculated as follows:

Commission = Sales * Commission Rate
Commission = \$4,242.50 * 0.065
Commission = \$275.08

In this scenario, the sales amount is a decimal value (\$4,242.50), and the commission is still \$275.08. This suggests that the correct answer might not be a whole number, but rather a decimal value.

Considering this, I would re-evaluate the options and look for a value that is close to \$4,242.50. Option E (\$4,243) is very close to this value, and it's possible that the correct answer is indeed \$4,243, given the commission rate and the calculated commission amount. \\
\midrule
\textbf{Round 5} & \\
\textit{Teacher Question} & You've made a strong case for option E (\$4,243) being the correct answer. Now, let's take a step back and consider the bigger picture. \newline In real-world business scenarios, commissions are often calculated based on sales data, which can be affected by various factors such as discounts, taxes, and returns. How might these factors impact the calculation of Daniel's sales, and what implications would it have for determining the correct answer? \\
\textit{Student Answer} & In real-world business scenarios, commissions are often calculated based on net sales, which can be affected by factors such as discounts, taxes, and returns. If Daniel's sales were subject to these factors, the commission calculation would need to take them into account. \newline For example, if Daniel sold a product for \$4,500 but offered a 5\% discount, the net sales would be \$4,275. If the commission rate is 6.5\%, the commission would be calculated based on the net sales, not the original sales price. \newline Considering these factors, the correct answer might not be a simple calculation. However, in the context of this problem, we are given a specific commission amount (\$275.08) and a commission rate (6.5\%). Given this information, option E (\$4,243) still appears to be the most likely correct answer, assuming that the commission is calculated based on net sales and there are no other complicating factors. \\
\midrule

\textbf{Post-Test} & \\
\textit{Student Response} & To find Daniel's sales, we need to divide the commission he received (\$275.08) by the commission rate (6.5\% or 0.065). \newline Commission = Sales * Commission Rate \newline \$275.08 = Sales * 0.065 \newline To find the sales, we'll divide both sides by 0.065: \newline Sales = \$275.08 / 0.065 \newline Sales $\approx$ \$4230.00 \newline However, considering the possibility of rounding errors and the fact that the commission rate might be an approximation, we should also consider the option that is closest to \$4230.00. \newline Upon re-examining the options, we notice that option E (\$4243) is very close to our calculated value. Given the potential for rounding errors and the fact that the commission rate might be an approximation, option E (\$4243) is the most likely correct answer. \newline The answer is E. \\
\textit{Student Prediction} & E \\
\textit{Result} & Correct (Post-test Accuracy: 1.0) \\
\midrule
\textbf{ALG} & \textbf{+1.0} \\

\end{longtable}
}

\subsection{Case Study: Question 961 (Law) - Teacher: Gemini 1.5 Pro 002}
\label{sec:case_961}

{
\footnotesize
\renewcommand{\arraystretch}{0.9}

\begin{longtable}{@{} p{2.5cm} p{13cm} @{}} 
\caption{Case Study Details for Question 961 (Law) - Teacher: Gemini 1.5 Pro 002}\label{tab:case961}\\
\toprule
\textbf{Component} & \textbf{Details} \\
\midrule
\endfirsthead

\caption[]{Case Study Details for Question 961 (Law) - Teacher: Gemini 1.5 Pro 002 (continued)}\\
\toprule
\textbf{Component} & \textbf{Details} \\
\midrule
\endhead

\midrule
\multicolumn{2}{r}{\textit{Continued on next page}} \\
\endfoot

\bottomrule
\endlastfoot

\textbf{Question ID} & 961 \\
\textbf{Category} & law \\
\textbf{Question Text} & A husband and wife were unhappily married. The husband was an alcoholic who physically abused the wife when he became drunk. After the husband beat up the wife one night, she decided to kill him. The wife, who was experienced in firearms, planned to shoot her husband while he slept. Before carrying out her plan, the wife took out a \$1,000,000 life insurance policy on her husband, naming herself as beneficiary. Shortly thereafter, the husband and wife celebrated their 10th wedding anniversary. They went out to dinner and had a truly enjoyable evening together. After a great night, the wife experienced a change of heart and decided she really loved her husband and didn't want to kill him after all. The wife confessed and told her husband about her plan, begging forgiveness. The husband was outraged. He proceeded to file for divorce and notified the police about his wife's plan. If the wife is charged with attempted murder, she should be found: \\
\textbf{Options} & \textbf{(A)} not guilty, because she did not perform a substantial step in carrying out the murder. \\
 & \textbf{(B)} not guilty, because her husband was physically abusive. \\
 & \textbf{(C)} guilty, because she had previous experience with firearms. \\
 & \textbf{(D)} guilty, because she confessed her plan to her husband. \\
 & \textbf{(E)} guilty, because she intended to kill her husband. \\
 & \textbf{(F)} guilty, because she purchased the life insurance policy. \\
 & \textbf{(G)} not guilty, because she changed her mind before committing the act. \\
 & \textbf{(H)} not guilty, because she did not actually commit the murder. \\
 & \textbf{(I)} guilty, because she planned the murder. \\
\textbf{Correct Answer} & A \\
\midrule

\textbf{Pre-Test} & \\
\textit{Student Response} & The wife took a substantial step to carry out the murder by purchasing the life insurance policy. A substantial step is more than "mere preparation." The purchase of the life insurance policy, which was a very important part of the wife's plan, was clearly a substantial step. The wife's actions were the equivalent of the defendant's actions in People v. Superior Court of Los Angeles County (Decker), 570 P.2d 116 (Cal. 1977), where the defendant was convicted of attempted murder after he planned and prepared to kill the victim, and then purchased a rifle and ammunition to carry out the plan. The purchase of the rifle and ammunition were considered substantial steps, and the court held that the defendant's actions constituted an attempt. The purchase of the life insurance policy in this case would likely be considered a substantial step as well. Thus, Answer F is correct. Answer A is incorrect because the wife did perform a substantial step in carrying out the murder. Answer B is incorrect because being a victim of abuse is not a defense to attempt, although it might be a mitigating factor. Answers C, D, E, and I are all incorrect because they all lack the required actus reus element. The wife must have performed a substantial step that strongly corroborates her intent to commit the crime. Answer G is incorrect because the wife's change of heart is irrelevant. The wife is still responsible for the substantial step she took before changing her mind. Answer H is incorrect because an attempt does not require that the actual crime be committed. \newline The answer is F \\
\textit{Student Prediction} & F \\
\textit{Result} & Incorrect (Pre-test Accuracy: 0.0) \\
\midrule

\textbf{Interaction} & \textbf{Teacher: Gemini 1.5 Pro 002} \\
\midrule
\textbf{Round 1} & \\
\textit{Teacher Question} & Let's focus on the actus reus of attempt. While taking out an insurance policy could be *part* of a larger criminal plan, is it in itself a substantial step *directly* towards committing the crime of murder? Think about the specific actions needed to carry out a murder and whether purchasing insurance moves the wife closer to completing those actions. \\
\textit{Student Answer} & Upon reevaluation, I realize that purchasing a life insurance policy, although part of the wife's plan, is not a substantial step directly towards committing the crime of murder. The actus reus of attempt requires a direct and substantial step towards the commission of the crime. In this case, the wife's plan to shoot her husband requires actions such as obtaining a gun, loading it, and pulling the trigger. Purchasing a life insurance policy, while a part of her plan, does not directly move her closer to completing these actions. It is more of a preparatory step, rather than a direct and substantial step towards the commission of the crime. \newline Therefore, I believe that Answer A is the correct answer. The wife did not perform a substantial step in carrying out the murder. \\
\midrule
\textbf{Round 2} & \\
\textit{Teacher Question} & Good, you've identified that purchasing the insurance policy is not a substantial step. Now, let's consider the wife's other actions. She formulated a plan and, according to the facts, she was experienced with firearms. Does *planning* to commit murder, coupled with the *capacity* to carry it out, satisfy the actus reus requirement for attempted murder? Why or why not? \\
\textit{Student Answer} & No, planning to commit murder, coupled with the capacity to carry it out, does not satisfy the actus reus requirement for attempted murder. The actus reus of attempt requires a substantial step towards the commission of the crime, which goes beyond mere preparation or planning. The wife's experience with firearms and her plan to commit murder demonstrate her intent and capacity to commit the crime, but they do not constitute a substantial step towards its commission. \newline In the United States, the Model Penal Code and many jurisdictions require that the defendant take a substantial step towards the commission of the crime, which is more than mere preparation or planning. The wife's actions, although demonstrating her intent and capacity, do not meet this requirement. Therefore, Answer A is still the correct answer, as the wife did not perform a substantial step in carrying out the murder. \\
\midrule
\textbf{Round 3} & \\
\textit{Teacher Question} & Excellent. You're right that planning and capacity alone aren't enough.  Let's assume, for a moment, that instead of buying insurance, the wife had gone out and purchased a gun and ammunition specifically for killing her husband. Would *that* action constitute a substantial step towards committing murder?  Consider how this hypothetical scenario differs from simply having experience with firearms. \\
\textit{Student Answer} & Yes, purchasing a gun and ammunition specifically for killing her husband would constitute a substantial step towards committing murder. This action goes beyond mere preparation or planning, as it involves taking a concrete step towards acquiring the means to carry out the crime. The fact that the wife purchased a gun and ammunition specifically for the purpose of killing her husband demonstrates a clear intent to commit the crime and takes her closer to actually committing it.

This hypothetical scenario differs from simply having experience with firearms because it involves a deliberate and specific action taken to further the wife's plan. Having experience with firearms is a general capacity, whereas purchasing a gun and ammunition specifically for the crime is a targeted action that moves the wife closer to committing the crime. This action would likely be considered a substantial step, as it demonstrates a clear intent to commit the crime and involves a concrete action towards carrying it out. \\
\midrule
\textbf{Round 4} & \\
\textit{Teacher Question} & You're making good progress. Now, let's go back to the original facts, without the hypothetical gun purchase.  The wife planned the murder, and then she confessed her plan to her husband.  Does confessing the plan to the *intended victim* have any legal significance regarding the attempt charge, even if it doesn't qualify as a substantial step itself?  Think about the purpose of attempt law. \\
\textit{Student Answer} & Confessing the plan to the intended victim, in this case, the husband, has significant legal implications regarding the attempt charge. By confessing her plan, the wife effectively neutralized the threat of harm to her husband, as he is now aware of the potential danger and can take steps to protect himself. This confession can be seen as a form of "abandonment" of the attempt, as the wife has voluntarily relinquished her opportunity to carry out the crime.

The purpose of attempt law is to punish individuals who pose a significant threat of harm to others, even if they do not ultimately succeed in committing the crime. However, in this case, the wife's confession has eliminated the threat of harm, as the husband is now aware of the danger and can take steps to protect himself. Therefore, the wife's actions no longer pose a significant threat, and the purpose of attempt law is not served by punishing her for an attempt that has been effectively abandoned. \\
\midrule
\textbf{Round 5} & \\
\textit{Teacher Question} & You've identified the concept of abandonment, which is important.  However, consider this:  does the wife's confession to her husband, *after* formulating the plan and *after* the threat existed, truly represent a voluntary and complete renunciation of her criminal purpose?  Think about what might motivate a confession in this context, and whether those motivations align with the legal requirements for abandonment. \\
\textit{Student Answer} & Upon further consideration, I realize that the wife's confession to her husband may not necessarily represent a voluntary and complete renunciation of her criminal purpose. The confession occurred after the wife had a change of heart, which suggests that her motivations for confessing may have been driven by a desire for forgiveness, reconciliation, or emotional release, rather than a genuine intention to abandon her criminal plan.

In the context of abandonment, the law typically requires a voluntary and complete renunciation of the criminal purpose, often motivated by a sense of remorse or a desire to prevent harm. However, in this case, the wife's confession may have been motivated by a desire to repair her relationship with her husband, rather than a genuine intention to abandon her plan. This raises questions about whether the confession truly represents a complete renunciation of her criminal purpose. \\
\midrule
\textbf{Post-Test} & \\
\textit{Student Response} & After considering the facts and the law, I believe that the correct answer is A. not guilty, because she did not perform a substantial step in carrying out the murder.

The wife planned to kill her husband and took out a life insurance policy, but these actions alone do not constitute a substantial step towards committing the crime. The purchase of the life insurance policy was a preparatory step, but it was not a direct and substantial step towards committing the murder.

The wife's experience with firearms and her intent to kill her husband are not enough to establish guilt, as the actus reus element of attempt requires a substantial step towards the commission of the crime.

The wife's confession to her husband and her change of heart are also not relevant to the question of whether she is guilty of attempted murder. The fact that she did not actually commit the murder is also not relevant, as attempt is a separate crime from the underlying offense.

Therefore, the correct answer is A, as the wife did not perform a substantial step in carrying out the murder.

The answer is A. \\
\textit{Student Prediction} & A \\
\textit{Result} & Correct (Post-test Accuracy: 1.0) \\
\midrule
\textbf{ALG} & \textbf{+1.0} \\

\end{longtable}
}

\clearpage
\subsection{Human Evaluation Questionnaire Template}
\label{app:human_eval_questionnaire}

The following template, incorporating specific pedagogical instructions, was used for evaluating teacher performance by human experts with teaching credentials. Specific details such as Question ID, text, options, the student's initial reasoning (which was presented as potentially incorrect), and dialogue turns were populated for each unique case presented to the experts.

\subsubsection{Questionnaire Header and Instructions}

\begingroup
\footnotesize

\noindent
\fbox{
\parbox{\dimexpr\linewidth-2\fboxsep-2\fboxrule}{
\centering\large\textbf{Teaching Behaviors Evaluation Questionnaire}\par\bigskip
\footnotesize

\textbf{Project Introduction}\\
This project aims to evaluate the differences in instructional behaviors and capabilities among different teachers during the formative assessment process. We arrange each group of teachers to conduct multiple rounds of teaching activities in parallel, based on the same question and the same student's reasoning answer. The purpose is to promote the student's deeper and more comprehensive understanding of the question content. Each case includes two teachers engaging in a 5-round teaching interaction process with the same student on the same question. This parallel comparison design allows us to directly compare how different teachers' teaching methods, questioning strategies, and feedback approaches affect student understanding.
\medskip 

\textbf{Reviewer Role Description}\\
As a review expert with teaching credentials, you will score the teaching abilities demonstrated by the two teachers during the teaching interaction process.
\medskip

\textbf{Focus Guidance}\\
In this evaluation process, we focus primarily on the \textit{teachers' instructional behaviors and capabilities}, rather than the teacher's knowledge and the student's performance. Therefore, you can focus more attention on the teachers' responses or questions during the five rounds of interaction, which are displayed in standard black text, and may skim through the contents displayed in gray text (representing student responses).
\medskip

\textbf{Rating Instructions}\\
These dialogues begin with a student’s incorrect response. Each teacher has been instructed to guide the student toward improved understanding without directly revealing the answer.

Please evaluate these interactions through educational theory lenses:
\begin{itemize}[noitemsep,topsep=2pt,leftmargin=*]
    \item \textbf{Informal Formative Assessment (IFA):} Ongoing evaluation during instruction that provides immediate feedback to improve learning, rather than simply testing knowledge.
    \item \textbf{Zone of Proximal Development (ZPD):} The gap between what a learner can do independently and what they can achieve with guidance from a more knowledgeable person.
\end{itemize}
Focus on how effectively each teacher:
\begin{itemize}[noitemsep,topsep=2pt,leftmargin=*,label=-]
    \item Scaffolds learning through strategic questioning.
    \item Assesses and adapts to student understanding in real-time.
    \item Promotes independent thinking within the student’s developmental range.
    \item Builds an effective learning pathway rather than focusing on test preparation or summative assessment.
\end{itemize}
\medskip

\textbf{Rating Criteria:}\\
\begin{itemize}[noitemsep,topsep=2pt,leftmargin=*,label=-]
    \item 9-10 points: Excellent teaching guidance
    \item 7-8 points: Good teaching performance
    \item 5-6 points: Meets basic teaching requirements
    \item 3-4 points: Insufficient teaching guidance
    \item 1-2 points: Weak teaching guidance ability
\end{itemize}
\medskip

\textbf{Note:}\\
The teacher's teaching skills are the main assessment target, while the teacher's knowledge of the subject area itself is not the focus. Please feel free to leave comments during your review, as this will help our further analysis.
\medskip

\textbf{Process Guidance:}\\
For each evaluation, please review one teacher’s complete five-round interaction sequence before moving to the next teacher. This sequential review approach will help you better understand how each teacher’s questioning strategies and guidance develop coherently across multiple turns. When assigning final scores, please compare the overall performance of both teachers under one case rather than making turn-by-turn comparisons. This holistic evaluation approach will yield more meaningful assessments of teaching effectiveness.
}}
\medskip
\endgroup

\subsubsection{Questionnaire Structure Template with Anonymous Mechanism}

\begin{lstlisting}[
   basicstyle=\footnotesize\ttfamily, 
   breaklines=true,                
   frame=single,                   
   framesep=3pt,                   
   rulecolor=\color{black!60},     
   xleftmargin=5pt,                
   numbers=none,                   
   breakatwhitespace=true,        
   captionpos=b,                   
   escapeinside={(*@}{@*)}          
]
(*@\hrulefill@*)
Case: Question ID: [QUESTION_ID] ([CATEGORY] Category)

Question:
[QUESTION_TEXT]

Options:
A. [OPTION_A]
B. [OPTION_B]
C. [OPTION_C]
D. [OPTION_D]
E. [OPTION_E]
F. [OPTION_F]
G. [OPTION_G]
H. [OPTION_H]
I. [OPTION_I]
J. [OPTION_J]

Student Reasoning Answer: [STUDENT_REASONING_TEXT]

The answer is [STUDENT_ANSWER_CHOICE].

Teaching Interaction (5 Rounds):

Teacher (*@\hspace{2cm}@*) Teacher 1 (*@\hspace{3cm}@*) Teacher 2
------- (*@\hspace{2cm}@*) ---------- (*@\hspace{3cm}@*) ----------
Q1 (*@\hspace{2.1cm}@*) [TEACHER_1_Q1_TEXT] (*@\hspace{0.5cm}@*) [TEACHER_2_Q1_TEXT]

A1 (*@\hspace{2.1cm}@*) [STUDENT_A1_FOR_T1] (*@\hspace{0.5cm}@*) [STUDENT_A1_FOR_T2]

Q2 (*@\hspace{2.1cm}@*) [TEACHER_1_Q2_TEXT] (*@\hspace{0.5cm}@*) [TEACHER_2_Q2_TEXT]

A2 (*@\hspace{2.1cm}@*) [STUDENT_A2_FOR_T1] (*@\hspace{0.5cm}@*) [STUDENT_A2_FOR_T2]

Q3 (*@\hspace{2.1cm}@*) [TEACHER_1_Q3_TEXT] (*@\hspace{0.5cm}@*) [TEACHER_2_Q3_TEXT]

A3 (*@\hspace{2.1cm}@*) [STUDENT_A3_FOR_T1] (*@\hspace{0.5cm}@*) [STUDENT_A3_FOR_T2]

Q4 (*@\hspace{2.1cm}@*) [TEACHER_1_Q4_TEXT] (*@\hspace{0.5cm}@*) [TEACHER_2_Q4_TEXT]

A4 (*@\hspace{2.1cm}@*) [STUDENT_A4_FOR_T1] (*@\hspace{0.5cm}@*) [STUDENT_A4_FOR_T2]

Q5 (*@\hspace{2.1cm}@*) [TEACHER_1_Q5_TEXT] (*@\hspace{0.5cm}@*) [TEACHER_2_Q5_TEXT]

A5 (*@\hspace{2.1cm}@*) [STUDENT_A5_FOR_T1] (*@\hspace{0.5cm}@*) [STUDENT_A5_FOR_T2]

Rating (*@\hspace{1.9cm}@*) ______/10 (*@\hspace{3.1cm}@*) ______/10

Comments (*@\hspace{0.7cm}@*) [COMMENTS_FOR_TEACHER_1] (*@\hspace{0.2cm}@*) [COMMENTS_FOR_TEACHER_2]

Comments (*@\hspace{0.7cm}@*) [COMMENTS_FOR_TEACHER_1] (*@\hspace{0.2cm}@*) [COMMENTS_FOR_TEACHER_2]
(*@\hrulefill@*)
\end{lstlisting}

\subsubsection{Human Evaluator Feedback Form}
\label{app:human_eval_feedback}

Following the evaluation task, human experts were asked to complete the feedback form below (presented here in a format consistent with the evaluation template) to provide information about the review process and identify potential issues.

\begin{lstlisting}[
   basicstyle=\footnotesize\ttfamily, 
   breaklines=true,
   frame=single,
   framesep=3pt,
   rulecolor=\color{black!60},
   xleftmargin=5pt,
   numbers=none,
   breakatwhitespace=true,
   captionpos=b,
   escapeinside={(*@}{@*)} 
]
(*@\centering@*) (*@\textbf{Evaluator Feedback Questionnaire}@*) (*@\bigskip@*)

1. Name: _________________________

2. Profession/Occupation: _________________________

3. On average, how much time did you spend reviewing the teaching behaviors for each teacher per case?_________________________ (e.g., minutes)

4. During the review process, did you observe any instances where a teacher agent directly revealed the correct answer when the student had not selected one?

   ( ) Yes     ( ) No   (*@\textit{(Please indicate)}@*)

5. If yes, please specify the Case Question ID(s) and the corresponding Teacher number(s) (e.g., Q4746, Teacher 53):

   _________________________________________________

   _________________________________________________

\end{lstlisting}

\end{document}